\pdfoutput=1
\documentclass[11pt]{article}

\usepackage[preprint]{acl}

\usepackage{float}
\usepackage{times}
\usepackage{latexsym}
\usepackage{amssymb}
\usepackage{colortbl}
\usepackage[table,xcdraw]{xcolor}
\usepackage{subcaption}
\usepackage{listings}
\usepackage[T1]{fontenc}
\lstdefinestyle{compactcode}{
  basicstyle=\ttfamily\small,
  columns=fullflexible,
  upquote=true,
  showstringspaces=false,
  keepspaces=true,
  breaklines=true,
  aboveskip=4pt,
  belowskip=4pt,
  xleftmargin=0pt,
  frame=single,
  framesep=2pt,
  rulecolor=\color{black},
}
\usepackage{enumitem}     % For compact lists

\usepackage[most]{tcolorbox}  % Required for "enhanced", "breakable", "skins"
\usepackage{fontawesome5}     % Required for icons like \faRobot, \faCogs

\definecolor{boxframe}{RGB}{40, 44, 52}       % Dark Grey (System)
\definecolor{boxback}{RGB}{248, 249, 251}     % Off-White (System)
\definecolor{keycolor}{RGB}{200, 0, 0}        % Red for JSON Keys
\definecolor{keywordblue}{RGB}{0, 83, 156}    % Royal Blue (Fixed: was missing)

\definecolor{method_blue}{RGB}{0, 122, 255}   % Azure Blue
\definecolor{method_purple}{RGB}{120, 80, 220}% Deep Purple (CoT)
\definecolor{method_teal}{RGB}{0, 150, 136}   % Teal (Pipelines)
\definecolor{method_emerald}{RGB}{16, 185, 129}% Emerald (T3)
\definecolor{method_orange}{RGB}{234, 88, 12} % Orange (Rules)

\newtcolorbox{promptbox}[2][]{
  enhanced,
  breakable,
  title={\faRobot\ \ \textbf{#2}},
  colframe=boxframe,
  colback=boxback,
  coltitle=white,
  fonttitle=\small\sffamily,
  fontupper=\small\ttfamily,
  boxrule=0.5mm,
  sharp corners,
  drop shadow,  % Fixed: Changed from "medium" to generic for compatibility
  left=10pt, right=10pt, top=10pt, bottom=10pt,
  #1
}

\newtcolorbox{methodbox}[3][]{
  enhanced,
  breakable,
  skin=bicolor,
  colback=white,
  colbacktitle=white,
  colframe=gray!20,
  coltitle=black,
  attach boxed title to top left={xshift=0mm, yshift=0mm},
  boxed title style={frame hidden, size=small, bottom=0mm},
  borderline west={4pt}{0pt}{#2}, % Colored Sidebar
  title={%
    \textcolor{#2}{\faCogs}\ \ \textbf{\small \MakeUppercase{#3}}%
  },
  fontupper=\small\ttfamily,
  left=12pt, right=10pt, top=5pt, bottom=10pt,
  sharp corners,
  drop shadow,  % Fixed: generic shadow
  #1
}

\newcommand{\jsonkey}[1]{\textbf{\textcolor{keycolor}{#1}}}

\lstdefinelanguage{promptlang}{
  keywords={Input, Output, Instruction, Role, Batsman, Bowler, Rules, Format},
  keywordstyle=\color{keywordblue}\bfseries,
  basicstyle=\ttfamily\small\color{black!80},
  breaklines=true,
  frame=none
}
\usepackage[utf8]{inputenc}

\usepackage{microtype}

\usepackage{inconsolata}

\usepackage{graphicx}
\usepackage{booktabs}
\usepackage{rotating}
\usepackage{pgfplotstable}
\usepgfplotslibrary{groupplots}
\pgfplotsset{compat=1.18}
\usepackage{pifont}

\usepackage{soul}  % for \hl{} highlighting
\sethlcolor{yellow}
\usepackage{makecell}
\usepackage{textcomp}
\usepackage{amsfonts}
\usepackage{multirow}
\usepackage{tabularx}
\newcolumntype{Y}{>{\centering\arraybackslash}X}
\definecolor{negred}{HTML}{FD6864}
\definecolor{posgreen}{HTML}{32CB00}
\definecolor{rowgray}{HTML}{EFEFEF}
\definecolor{headergray}{HTML}{C0C0C0}
\definecolor{headercolor}{RGB}{230, 240, 250} % Light blue-ish gray
\definecolor{rowcolor}{RGB}{250, 250, 250}    % Very light gray
\title{Moneyball with LLMs:
Analyzing Tabular Summarization in Sports Narratives}

\author{
\textsuperscript {} Ritam Upadhyay\textsuperscript{1} \thanks{These authors contributed equally} \quad
\textsuperscript {} Naman Ahuja\textsuperscript{1} \footnotemark[1] \quad
\textsuperscript {} Rishabh Baral \textsuperscript{1} \quad
\textsuperscript{}Aparna Garimella \textsuperscript{2} \quad
\textsuperscript {} Vivek Gupta \thanks{Primary supervisor for this work}\textsuperscript{1} \\ 
\\ 
\textsuperscript{1}School of Computing and Augmented Intelligence, Arizona State University \\
\textsuperscript{2}Adobe Research, India \\
\\
\texttt{rbaral1@asu.edu}, \texttt{rupadh17@asu.edu}, \texttt{nahuja11@asu.edu}, \texttt{garimell@adobe.com},\texttt{\textsuperscript{\dag}vgupt140@asu.edu}
}

\begin{document}
\definecolor{negred}{HTML}{FD6864}
\definecolor{posgreen}{HTML}{32CB00}
\definecolor{colorOrig}{RGB}{31, 119, 180}   % Blue
\definecolor{colorCount}{RGB}{255, 127, 14}  % Orange
\definecolor{colorPara}{RGB}{44, 160, 44}    % Green
\definecolor{colorAnon}{RGB}{214, 39, 40}    % Red
\maketitle
\begin{abstract}
% Large language model (LLM) approaches to tabular summarization often rely on extensive prompt engineering, decomposition pipelines, or entity-level intermediate representations to achieve strong performance. While these strategies improve accuracy, they are computationally expensive and provide limited insight into how effectively models maintain state over long, evolving narratives. In this work, we introduce \textsc{SporTabSet}, a diagnostic benchmark for long-context tabular summarization spanning two complementary sports domains that require tracking multiple entities and aggregating statistics under domain-specific rules. Using SporTabSet, we conduct a systematic evaluation of decomposition-based modeling strategies across several long-context LLMs. Our results show that while decomposition substantially improves accuracy and numerical fidelity, gains are driven primarily by reducing multi-entity interference rather than improved local arithmetic. We further demonstrate through robustness experiments that models remain highly sensitive to surface-level cues such as entity forms and summary patterns, exhibiting structured failure modes including hallucination, omission, and role confusion. Together, these findings indicate that maintaining consistent multi entity memory is a central bottleneck in long context table generation, motivating diagnostic evaluation as a prerequisite for developing scalable and reliable foundational systems for tabular summarization.
Large language model (LLM) approaches to tabular summarization rely on extensive prompt engineering, decomposition pipelines, or entity-level intermediate representations to achieve strong performance. While effective, these strategies are computationally expensive and offer limited insight into how well models maintain state over long, evolving narratives. We introduce \textsc{SporTabSet}, a diagnostic benchmark for long-context tabular summarization across two complementary sports domains that require tracking multiple entities and aggregating statistics under domain-specific rules. Using SporTabSet, we systematically evaluate decomposition-based strategies across several long context LLMs. Results show that although decomposition substantially improves accuracy and numerical fidelity, gains stem mainly from dissecting multi-entity interference rather than improved local arithmetic. Robustness experiments further reveal high sensitivity to surface-level cues with structured failures, including hallucination, omission, and role confusion. Together, these findings identify consistent multi-entity memory as a key bottleneck in long-context table generation, motivating diagnostic evaluation as a prerequisite for scalable, efficient and reliable tabular summarization models.

\end{abstract}

\section{Introduction}
\vspace{-0.35em}
% The web is a sprawling repository of human knowledge, much of it encoded in unstructured narratives such as blogs, social media posts, event streams, live commentary, and informal reports. As this content grows in volume and diversity, extracting structured knowledge <cite two papers called structsum here> from free-form text, also known as Text-to-Table generation, has gained importance.  While Large Language Models (LLMs) have shown promise in structured summarization tasks <cite text to table>, much of their progress has been measured on static benchmarks or narrow domains where schemas are fixed and context is limited. This leaves open the question of whether LLMs can remain robust under more dynamic, noisy, and state-dependent narratives.

\begin{figure}[t] 
    \centering
    \includegraphics[height=8cm]{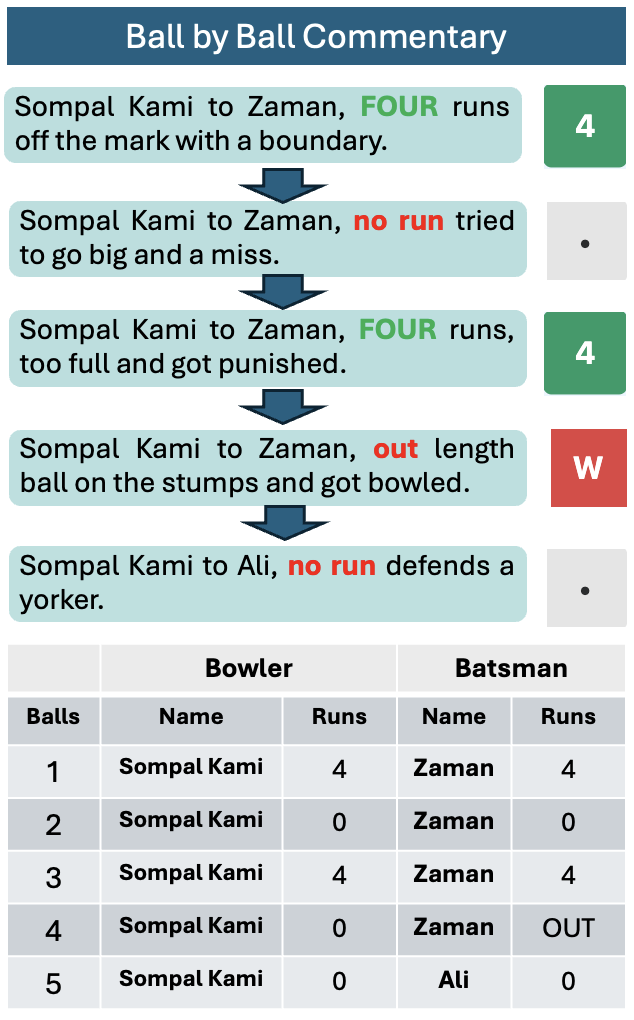}
    \vspace{-1.0em}
    \caption{\small Ball by Ball Commentary in Cricket and updation of Batsman and bowler Tables with every ball}
    \label{fig:commentary_flow}
    \vspace{-1.0em}
\end{figure}
Tables are a central abstraction for summarizing and reasoning over complex information, enabling compact storage, comparison, and verification across domains such as scientific reporting, finance, and sports analytics. Early work on inferring structured summaries from dense, unstructured text, commonly referred to as Text-to-Table Generation or Tabular Summarization, primarily framed the task as an information extraction problem, emphasizing structural integrity, schema alignment, and exact cell recovery from short or moderately sized inputs \cite{text-2-table-2022-acl,li2023sequence,sundar2024gtbls,pietruszka2024stable}.

More recent advances in large language models and structured generation interfaces, such as constrained decoding \cite{wang2025slotstructuringoutputlarge}, function calling, and tool-augmented agents \cite{schick2023toolformer}, have made it increasingly feasible to produce well-formed structured outputs directly from LLMs \cite{hu2025rlstructlightweightreinforcementlearning,li-etal-2025-dice,yao2022react}. This shift has enabled approaches that go beyond surface extraction, incorporating temporal reasoning over long contexts to generate summarized representations from evolving narratives \cite{hu2024sportsmetricsblendingtextnumerical}. However, sustaining such behavior requires models to maintain evolving states of multiple attributes across many interacting entities, a challenge widely recognized in long-context and memory-augmented language modeling \cite{wang2023augmenting,fang2024unimem}.

To mitigate this, recent methods externalize memory through iterative prompting \cite{ahuja-etal-2025-map}, structured memory representations \cite{deng-etal-2024-text}, or recursive summarization schemes \cite{wang2025recursivelysummarizingenableslongterm}. While effective for tabular summarization, these designs often obscure how faithfully models maintain internal state over long contexts. Moreover, most table generation frameworks are evaluated only \textbf{one} dynamic benchmark (LiveSum)\cite{deng-etal-2024-text}, providing limited insight into how different modeling strategies generalize across narrative structures, table sizes, and entity regimes \cite{hu2024sportsmetricsblendingtextnumerical}. As a result, it remains unclear \emph{whether LLMs support long-context, multi-entity reasoning, which relies on surface regularities, and what trade-offs they incur in terms of accuracy, robustness, and computational cost}. This question is particularly compelling for LLMs, as they offer a unified interface for reasoning, memory, and generation, but lack explicit mechanisms for maintaining persistent, structured state.

To address these limitations, we introduce \textsc{SporTabSet}, a diagnostic benchmark designed to systematically evaluate modeling strategies for dynamic text-to-table generation under long-context, multi-entity settings. SporTabSet spans two complementary sports domains: cricket and basketball, that expose distinct but related challenges in tabular summarization. Cricket requires maintaining and aggregating interdependent statistics across multiple evolving tables governed by explicit role-based game rules (batsmen and bowlers), while basketball emphasizes wide table construction with locally attributed updates across frequent entities. Together, these domains enable controlled analysis of how different modeling strategies handle entity tracking, temporal aggregation, and numerical consistency beyond short-context or purely extractive settings.

Our contributions are threefold:
\begin{itemize}
    \item We introduce \textsc{SporTabSet}, a multi-sport benchmark for long-context text-to-table generation, comprising over 6,500 cricket commentaries with programmatically derived ground truth and 2500 basketball games aligned with official box scores, along with a suite of label-invariant perturbations.
    \item We conduct a systematic comparison of modeling strategies for dynamic tabular summarization, from standard CoT \cite{cot} prompting to decomposition-based methods, entity-centric generation, and symbolic pipelines, analyzing their trade-offs across accuracy, numerical fidelity, and practical implications.
    \vspace{-0.5em}
    \item We present a entity-centric robustness analysis that reveals structured failure modes such as hallucination, omission, role-confusion providing empirical evidence that maintaining multi-entity state is a primary bottleneck in long-context table generation with LLMs.
\end{itemize}

% \begin{table*}[h]
% \centering
% \small
% \begin{tabular}{@{}lcccc@{}}
% \toprule
% \textbf{Dataset} & \textbf{Task}& \textbf{Multi-table} & \textbf{Dynamic Schema}& \textbf{Avg. Tokens}\\ 
% \midrule
% RotoWire & Extraction & \cmark & \cmark& 347 \\ 
% WikiBio & Extraction & \xmark & \xmark& 95 \\ 
% WikiTableText & Extraction & \xmark & \xmark& 14 \\ 
% E2E & Extraction & \xmark & \xmark& 27 \\ 
% Livesum & Counting & \xmark & \xmark& 1,571 \\ 
% SportsMetrics & Count. + Reas. & \xmark & \cmark& 6,229 / 6,166 \\ 
% \textbf{CMT-Bench} & \textbf{Count. + Reas. + Extr.} & \cmark & \cmark& \textbf{9435} \\ 
% \bottomrule
% \end{tabular}
% \caption{Comparison of text-to-table benchmarks. Extraction refers to tasks where values from input text are directly placed in tables. Counting refers to tasks that require counting the number of events that uniformly increment by 1, eg, Wickets, Balls. Reasoning refers to tasks that require contextual understanding, for eg, A Maiden Over refers to consecutive six deliveries by a bowler giving 0 runs.}
% \label{tab:dataset_comparison}
% \end{table*}

\section{SporTabSet Benchmark}
Cricket and basketball together form a complementary, entity-driven benchmark for evaluating large language models on dynamic text-to-table generation. In both sports, the task centers on resolving implicit references, performing numerical reasoning, and making consistent updates under a complex domain policy. This presents complementary challenges from previous SoTA benchmarks in this domain \cite{deng-etal-2024-text} where each event induces only singular updates to a fixed table schema across the benchmark. 

Despite these shared properties, the two sports impose distinct structural demands. Cricket presents long-horizon, multi-entity interactions with two distinct yet interdependent roles i.e. batsmen and bowlers, each requiring separate output tables and different logical reasoning: batsmen participate continuously across consecutive segments of play, while bowlers contribute episodically in distinct phases, necessitating phase-aware reasoning. Basketball commentary, in contrast, involves many agents operating within a single functional role, typically requiring the generation of one consolidated table covering a large roster of players (on average 24), where reasoning is dominated by dense, overlapping interactions occurring simultaneously rather than by role-specific or phase-structured temporal dependencies. With cricket stressing multi-table coordination and basketball requiring wide table construction, these datasets enable a systematic evaluation of LLM capabilities and robustness across diverse narrative and statistical regimes.

\paragraph{Dataset Collection and Preprocessing.}
% \label{sec:dataset_construction}

We collect commentary–scorecard pairs from two sports domains: cricket from \textbf{Cricinfo}\footnote{\url{https://www.espncricinfo.com/}} and basketball from \textbf{ESPN}\footnote{\url{https://www.espn.com/nba/scoreboard}}. For cricket, we use ball-by-ball commentary covering complete innings. To ensure chronological integrity and remove non-play content (e.g., weather updates, advertisements), we apply a monotonicity filter based on two criteria: \textbf{(i) Entity presence:} both the bowler and batsman names must co-occur in the line, separated by the standard delimiter pattern used across commentaries; and \textbf{(ii) Action presence:} the line must include a valid gameplay outcome such as \textit{no run}, \textit{FOUR}, \textit{OUT}, or an extra (e.g., \textit{1w}, \textit{1nb}). Our base dataset comprises \textbf{498 ODI innings (2006–2025)}\footnote{One Day International of approximately 300 balls, played between national teams}. 

To analyze model performance across varying context lengths, we additionally create truncated samples by slicing the first $x$ (120/180/240) balls of randomly selected samples, enabling systematic evaluation of temporal robustness and the effect of input length on model consistency. Each commentary is paired with two gold tables, a batsman scorecard and a bowler scorecard. 

For basketball, we collect commentary for 998 games from the 2025 season, sourced from ESPN. Each commentary is paired with the corresponding official final box score, which is scraped and used as the ground-truth table. Basketball commentary does not permit deterministic reconstruction of all statistics from text alone; therefore, the box score provides verified per-player statistics such as points, rebounds, and assists. All data are sourced from publicly available pages under fair-use terms for academic research. A sample commentary and corresponding ground-truth tables are provided in Appendix \ref{sec: example_gnd_truth}.

\paragraph{Ground Truth Creation.}
% \label{sec:gt_creation}

% \begin{itemize}
%     \item \textbf{Batsman Tables} include \textit{runs}, \textit{balls faced}, \textit{fours}, \textit{sixes}, and \textit{strike rate (S/R)}. These attributes require both counting (e.g., runs and boundaries) and arithmetic reasoning (e.g., strike rate = runs/balls). Models must also handle figurative boundary expressions and implicit event cues within commentary text.
    
%     \item \textbf{Bowler Tables} include \textit{balls}, \textit{runs conceded}, \textit{wickets}, \textit{overs}, and \textit{maidens}. Bowling statistics require cross-span reasoning, e.g., identifying a maiden over demands tracking six consecutive deliveries with zero runs, and careful attribution of extras (wides, no-balls, byes)
% \end{itemize}

To generate the ground truth tables for truncated samples, that are not publicly available online, each innings is segmented into ball-level events using regular expressions based on Cricinfo’s standard notation (e.g., \textit{3.2: Starc to Kohli}). A domain-specific parser extracts outcomes such as runs, boundaries, wickets, and extras. Cricket extras (\textit{wides, no-balls, byes, leg-byes}) are handled deterministically due to their asymmetric effects on batsman and bowler statistics, ensuring numeric consistency. Ball-level events are decomposed into constituent entities (Appendix~\ref{sec: Pseudocode}) and aggregated by player name to form the ground-truth batsman and bowler scorecards as shown in Table \ref{tab: Batsman Table Summary 5 overs} and Table \ref{tab: Bowler Table Summary 5 overs}.

For basketball, we construct a single player-level scorecard per game. Since commentary does not permit deterministic reconstruction of all statistics, we scrape the official final box score corresponding to each commentary to serve as ground truth. The table contains 10 numeric columns of standard per-player statistics (e.g., points, rebounds, assists etc.) as shown in Table \ref{tab: Basketball sample} and is aligned with commentary using game metadata. This setup evaluates a model’s ability to reconstruct a wide, entity-centric table from narrative text, even when statistics are only implicitly mentioned.

\paragraph{Ground-Truth Validation.}
Since cricket ground-truth tables are generated programmatically, we conduct human verification to ensure benchmark correctness. For samples containing complete match commentary, we compare the derived scorecards against publicly available official summaries and observe a 99.7\% exact match. For samples constructed via slicing, the authors manually validate 500 samples per setting, obtaining 99.3\% exact match at the cell level and 100\% accuracy for entity names (row headers).

For basketball, ground-truth tables are scraped directly from official sources. As the corresponding commentaries are not partially sliced, we perform manual verification on a random 10\% subset of the 998 games to confirm 100\% alignment between commentary and final box scores.

\section{Experiments}
% In our initial explorations cricket commentaries showed unusually high performance, with \textbf{GPT-4.1 achieving 89\% batsman cell accuracy}. Closer manual inspection revealed that commentaries contain \textbf{explicit batsman dismissal summaries} (e.g., \emph{run out (Chakabva) 5 (9b 0x4 0x6)}), which provide high-salience cues that models can exploit. To assess reliance on these cues, we mask all dismissal-related batsman summaries and re-evaluate performance, which causes a \textbf{significant drop in batsman accuracy across all models} (e.g., GPT-4.1: 89\% to 61\%, Gemini: 86\% to 49\%) (Table~\ref{tab:Masked_vs_unmasked}), indicating that batsman statistics are largely inferred from narrative summaries rather than ball-by-ball tracking. Interestingly, masking batsman cues also \textbf{degrades bowler performance}, suggesting that dismissal summaries act as shared structural anchors. 

% Open-weight models such as \textbf{Qwen-72B} and \textbf{LLaMA-70B} exhibit substantially higher sensitivity to cue removal, particularly in numeric error, revealing a tendency toward hallucinated precision when contextual anchors are weakened.
% We expand all the analysis for baselines and perturbations on the masked dataset of cricket removing all extractive cues. Basketball doesn't contain any such cues. In this section, we further describe the suite of LLMs benchmarked, the hybrid reasoning and decomposition-based baselines, and the evaluation metrics used to comprehensively evaluate performance on \textsc{SporTabSet}.

This section describes the evaluated LLMs, hybrid and decomposition-based baselines, and the metrics used to benchmark performance on \textsc{SporTabSet}.

\paragraph{Models.}
We evaluate state-of-the-art LLMs:  \textbf{GPT-4.1}~\cite{openai2024gpt4technicalreport}, \textbf{Gemini-2.5-Flash}, \textbf{Llama-3.3-70B-Instruct}~\cite{grattafiori2024llama3herdmodels} and, \textbf{Qwen-2.5-7B-Instruct}~\cite{qwen2025qwen25technicalreport}, and  Since cricket commentaries average approximately 9k tokens, we restrict our evaluation to models with context lengths exceeding 32k tokens and intruction tuned models adept at generating structured outputs. 

\paragraph{Baselines.}
Prior work~\cite{ball2024can} demonstrates that Zero-Shot Chain-of-Thought (ZS-CoT) prompting consistently outperforms few-shot prompting for table generation tasks. Additionally, decomposition-based approaches that scale test-time compute have been shown to yield substantial performance improvements~\cite{deng-etal-2024-text,ahuja-etal-2025-map}. Motivated by these findings, we evaluate ZS-CoT alongside a set of decomposition-based methods that combine LLM-generated representations with rule-based parsers, and analyze their performance relative to the computational cost incurred. Specifically, we implement the following methods:

\begin{enumerate}
    \item \textbf{ZS-CoT}: Models are provided with the cricket domain policy and instructed to reason step by step before producing the final scorecards.

    \item \textbf{Divide \& Generate \cite{jain-etal-2024-structsum}}: A two-stage pipeline in which the input commentary is partitioned into $n$ chunks. Each chunk is processed independently using ZS-CoT to generate intermediate scorecards in parallel. These intermediate outputs are then deterministically merged to produce the final scorecards. We experiment with $n \in \{2,4,8\}$.

    \item \textbf{Entity CoT}: This approach first prompts the LLM to extract the set of entities that defines the table schema. Each row is then generated using parallel LLM calls, with each call focusing on a single entity. While this increases input tokens (as each parallel call receives the full commentary), it simplifies the reasoning task by isolating entity-specific computation.

    \item \textbf{Text--Tuple--Table}: Following~\cite{deng-etal-2024-text}, this method first generates atomic tuples, one for each event of the form (Entity, Attribute, Value) signifying each update for the whole text. A rule-based compiler then aggregates these tuples to construct the final tables. This approach is the most computationally expensive, as generating atomic tuples over long-contexts significantly increases the number of output tokens per sample.
\end{enumerate}

\paragraph{Evaluation Metrics.}
We evaluate models using a combination of continuous and discrete metrics. Root Mean Squared Error (RMSE) and Symetric Mean Abs. Percentage Error (SMAPE) are used as quantitative metrics to measure numerical deviation between predicted and ground-truth table cells. SMAPE is particularly well-suited for commentary-driven scorecards, where many ground-truth cells are zero, as it avoids the disproportionate penalization of errors common with RMSE in sparse tables. In addition, we report Cell accuracy as a discrete correctness metric, measuring exact matches at the cell level to capture structural and symbolic fidelity of the generated tables.

A walk-through example for each method and the prompts used are provided in Appendix \ref{prompt}.

% \subsection{Evaluation Metrics}
% \label{eval_metrics}
% % Evaluating LLM-generated tables is notoriously hard because the same content can appear in many surface forms (merged columns, header reorderings, unit variants), making semantic matching metrics brittle \cite{Ahuja2025MapMake} \cite{tang-etal-2024-struc}. Recent work 
% State-of-the-art Table evaluation metrics  \cite{pancholi-etal-2025-tabxeval, ramu-etal-2024-bad, jain-etal-2024-structsum} primarily utilize LLM-driven frameworks that can capture structural fidelity and information loss beyond factual correctness. In our setting, however, the output tables are \emph{purely statistical} batsman and bowler scorecards with a \emph{fixed schema} and \emph{colorAnonical units} (counts for runs/balls/wickets), reducing correctness to numeric agreement under known constraints. We first align predicted and gold entities by normalizing names (lowercasing, whitespace removal) and computing a composite similarity (60\% token-overlap + 40\% \textit{SequenceMatcher}). We resolve the similarity matrix with the \textbf{Hungarian algorithm} and keep only alignments above $0.35$ to prevent spurious matches while tolerating minor noise. Post-alignment, we report \textbf{Cell Accuracy} (exact numeric agreement over all cells across both tables), \textbf{RMSE} (degree of error across all cells), and \textbf{SMAPE} (symmetric percentage error). 

% Add Metrics Tables here (@Ritam/@Naman...can you guys move it later?)
\begin{table*}[h!]
\small
\setlength{\aboverulesep}{0.35pt}
\setlength{\belowrulesep}{0.35pt}
\centering
% \renewcommand{\arraystretch}{1.2}
% Resize to fit text width
% \resizebox{\textwidth}{!}{%
\begin{tabular}{lccccccccc}
\toprule
\multirow{2}{*}{\textbf{Method}} & \multicolumn{3}{c}{\textbf{Cricket (Batsman)}} & \multicolumn{3}{c}{\textbf{Cricket (Bowler)}} & \multicolumn{3}{c}{\textbf{Basketball}} \\
\cmidrule(lr){2-4} \cmidrule(lr){5-7} \cmidrule(lr){8-10}
 & \textbf{Acc} & \textbf{RMSE} & \textbf{SMAPE} & \textbf{Acc} & \textbf{RMSE} & \textbf{SMAPE} & \textbf{Acc} & \textbf{RMSE} & \textbf{SMAPE} \\
\midrule

% --- BAND FOR MODEL NAME ---
\rowcolor[HTML]{C0C0C0} \multicolumn{10}{c}{\textbf{\textit{Llama-3.3 70B Instruct}}}\\

% --- DATA ROWS ---
CoT & 39 & 18.18 & 29 & 42 & 10.27 & 29 & 30 & 4.10 & 61 \\
Divide and Generate ($n=2$) & 40 & 15.19 & 21 & 40 & 12.34 & 31 & 33 & 6.15 & 57 \\
Divide and Generate ($n=4$) & 44 & 14.81 & 17 & 42 & 10.98 & \underline{25} & 47 & 4.66 & 36 \\
Divide and Generate ($n=8$) & \underline{48} & 12.93 & \underline{12} & \underline{45} & \underline{5.83} & \textbf{16} & \underline{58} & 2.84 & \underline{24} \\
EntityCoT & 41 & \underline{10.03} & 13 & 34 & 8.38 & \underline{25} & \textbf{62} & \textbf{1.55} & \textbf{19} \\
Text-Tuple-Table & \textbf{56} & \textbf{7.21} & \textbf{9} & \textbf{46} & \textbf{3.07} & \textbf{16} & 57 & \underline{1.69} & 27 \\

% --- BAND FOR MODEL NAME ---
\rowcolor[HTML]{C0C0C0} \multicolumn{10}{c}{\textbf{\textit{GPT-4.1}}}\\

% --- DATA ROWS ---
CoT & 53 & 9.58 & 15 & 47 & 7.34 & 24 & 47 & 1.70 & 36 \\
Divide and Generate ($n=2$) & 48 & 12.78 & 15 & 44 & 11.17 & 28 & 45 & 2.54 & 36 \\
Divide and Generate ($n=4$) & 48 & 13.71 & 13 & 48 & 8.38 & 21 & 50 & 2.16 & 30 \\
Divide and Generate ($n=8$) & 52 & 12.62 & \underline{10} & 49 & 4.22 & \underline{14} & 58 & 1.39 & 23 \\
EntityCoT & \underline{55} & \underline{7.14} & \underline{10} & \underline{57} & \underline{3.99} & \underline{14} & \underline{67} & \textbf{1.01} & \textbf{15} \\
Text-Tuple-Table & \textbf{64} & \textbf{4.23} & \textbf{5} & \textbf{59} & \textbf{2.15} & \textbf{12} & \textbf{68} & \underline{1.20} & \underline{17} \\

% --- BAND FOR MODEL NAME ---
\rowcolor[HTML]{C0C0C0} \multicolumn{10}{c}{\textbf{\textit{Gemini 2.5 Flash}}}\\

% --- DATA ROWS ---
CoT & 52 & 11.03 & 18 & 51 & 5.66 & 23 & 62 & 1.59 & 21 \\
Divide and Generate ($n=2$) & 81 & 3.39 & 3 & 84 & 1.11 & \underline{4} & 70 & 1.14 & 15 \\
Divide and Generate ($n=4$) & 83 & 3.02 & 3 & 86 & 1.04 & \underline{4} & \underline{74} & \underline{0.97} & \underline{13} \\
Divide and Generate ($n=8$) & 86 & 1.74 & \underline{2} & \underline{87} & \underline{0.86} & \underline{4} & \underline{74} & \textbf{0.96} & \underline{13} \\
EntityCoT & \underline{89} & \textbf{0.70} & \textbf{1} & 82 & 1.17 & \textbf{2} & 72 & 1.03 & \underline{13} \\
Text-Tuple-Table & \textbf{94} & \underline{0.96} & \textbf{1} & \textbf{89} & \textbf{0.59} & 6 & \textbf{75} & \underline{0.97} & \textbf{12} \\
\bottomrule
\end{tabular}
% }
\vspace{-0.75em}
\caption{Performance comparison on Cricket and Basketball datasets using Llama-3.3 70B Instruct, GPT-4.1, and Gemini 2.5 Flash}
\label{tab:unified_results}
\vspace{-2.0em}
\end{table*}

\subsection{Results and Analysis}
\label{sec:results}
In initial experiments, cricket commentaries exhibited unusually high performance, with \textbf{GPT-4.1 achieving 89\% batsman cell accuracy}. Manual inspection revealed that this performance is largely driven by \textbf{explicit batsman dismissal summaries} (e.g., \emph{run out (Chakabva) 5 (9b 0x4 0x6)}), which act as high-salience extractive cues. When all dismissal-related summaries are masked, accuracy drops sharply across models (e.g., GPT-4.1: 89\% to 61\%, Gemini: 86\% to 49\%; Table~\ref{tab:Masked_vs_unmasked}), indicating that batsman statistics are primarily inferred from narrative summaries rather than ball-by-ball reasoning. Notably, masking batsman cues also \textbf{degrades bowler accuracy}, suggesting that dismissal summaries function as shared structural anchors. 

We therefore perform all critical baseline and perturbation analyses on the masked cricket datasets, which removes direct extractive cues; basketball contains no analogous summaries. We first quantify the overall impact of task decomposition relative to a zero-shot Chain-of-Thought (CoT) baseline, and then compare decomposition strategies in detail. All reported numbers are derived from the unified results in Table~\ref{tab:unified_results}.

\paragraph{A. Overall Effect of Task Decomposition.}
% \label{sec:results_overall}

We begin by evaluating whether decomposing the tabular summarization task improves performance over a monolithic CoT baseline. Across all models, sports, and table types, we find that \emph{every} decomposition strategy yields substantial improvements in both accuracy and numerical fidelity.

Averaged across models and tasks, \textbf{Divide-and-Generate} with $n=8$ improves accuracy by \textbf{+14.9 pp} over CoT, while reducing RMSE by \textbf{37.8\%}. \textbf{EntityCoT} achieves a comparable average accuracy gain of \textbf{+15.1 points}, but with a larger \textbf{49.5\% reduction in RMSE}. The strongest improvements are obtained by \textbf{Text-Tuple-Table}, which yields an average accuracy gain of \textbf{+20.6 pp} and a \textbf{62.8\% reduction in RMSE}.

These gains are consistent across both sports. In cricket, decomposition improves both batsman and bowler tables, with mean accuracy gains ranging from approximately \textbf{+11 to +23 pp} depending on the method. In basketball, where tables are larger and entity density is higher, methods that reduce multi-entity interference achieve even larger gains, with EntityCoT improving accuracy by \textbf{+20.7 pp} on average.

\paragraph{B. Comparing Decomposition Strategies.}
% \label{sec:results_methods}
We now compare three classes of decomposition strategies: Divide and Generate (input chunking), EntityCoT (player-wise chunking), and Text-Tuple-Table (event-level chunking). While \textbf{Divide-and-Generate} improves performance with the number of segments $n$, the gains are uneven across models and sports. At $n=8$, Divide and Generate achieves a mean accuracy gain of \textbf{+14.9 pp} and a \textbf{37.8\% RMSE reduction} across tasks. The gains are strongest for \textbf{Gemini-2.5-Flash}, where cricket accuracy improves by \textbf{+34 pp} and basketball accuracy by \textbf{+12 pp} over CoT. \textbf{Llama-3.3} shows moderate but consistent gains, including improvements of up to \textbf{+28 pp} on basketball. In contrast, \textbf{GPT-4.1} exhibits unstable behavior at smaller $n$. 

% Notably, performance gains diminish or regress when intermediate scorecards are compiled using an LLM rather than rule-based aggregation (Table~\ref{tab:unified_results_2}). This highlights that while Divide-and-Generate reduces context length, \emph{table-level summation remains a fragile operation} for current models when handled generatively.

\textbf{EntityCoT} performs \emph{on par with or better than} Divide-and-Generate at $n=8$, despite requiring fewer decomposition steps. On average, EntityCoT improves accuracy by \textbf{+15.1 pp} and reduces RMSE by \textbf{49.5\%} relative to CoT. It is the best-performing method for basketball achieving a mean gain of \textbf{+20.7 pp}. RMSE reductions are particularly pronounced: on basketball, EntityCoT reduces RMSE by approximately \textbf{37--46\%} for GPT-4.1 and Llama-3.3, while Gemini-2.5-Flash achieves over \textbf{69\%} RMSE reduction. These results support our hypothesis that \textbf{multi-entity memory, rather than local arithmetic, is the primary bottleneck} in long-context table generation. Prior work on memory-augmented language models \cite{wang2023augmenting,fang2024unimem} has shown that retaining and retrieving long histories remains challenging even when local computation is straightforward. EntityCoT simplifies the reasoning problem and substantially reduces numerical drift, even when absolute accuracy gains are comparable to Divide-and-Generate.
Finally, \textbf{Text-Tuple-Table (T3)}delivers the strongest average performance across all methods. It achieves a mean accuracy gain of \textbf{+20.6 pp} and a \textbf{62.8\% reduction in RMSE}, ranking first on both metrics overall. T3 is the top-performing method for all three models in aggregate, with particularly large gains for \textbf{Gemini-2.5-Flash} (\textbf{+31 pp} accuracy, \textbf{73\%} RMSE reduction).

\paragraph{C. Model-Specific Behavior.}
% \label{sec:results_models}
We observe substantial variation in how models respond to decomposition strategies. \textbf{Gemini-2.5-Flash} shows the largest and most consistent gains across all methods, benefiting strongly from both decomposition and symbolic reasoning. \textbf{Llama-3.3} exhibits monotonic improvements as tasks are decomposed, particularly on basketball. In contrast, \textbf{GPT-4.1} shows inconsistent gains for Divide-and-Generate, while EntityCoT provides more reliable improvements. T3 exhibits high variance and severe reliability issues, including degenerate repetition and cumulative hallucination, especially for GPT-4.1 and Llama-3.3. A more detailed analysis of cost and reliability is given in the Appendix \ref{f_implementation details}. 

While these methods represent an upper bound on achievable performance, their practical reliability can be potentially limited without additional safeguards, underscoring the need for foundational models for this task.
\begin{table}[t!]
\small
\setlength{\aboverulesep}{0.25pt}
\setlength{\belowrulesep}{0.25pt}
\centering
\resizebox{\linewidth}{!}{%
\begin{tabular}{l|ccc|ccc}\toprule
\rowcolor{rowgray} & \multicolumn{3}{c|}{\textbf{Batsman}} & \multicolumn{3}{c}{\textbf{Bowler}} \\ \midrule
\rowcolor{rowgray} & \multicolumn{6}{c}{\textbf{Anonymization} (Original / Anonymized)}\\
\rowcolor{headergray} 
& Acc & RMSE & SMAPE & Acc & RMSE & SMAPE\\
\textbf{Gemini} & 52 / 53 & 11.0 / 13.1 & 18 / 20 & 51 / 55 & 5.6 / 4.7 & 23 / 18 \\
\textbf{2.5 Flash}& {\color{posgreen} +1} & {\color{negred} +2.09} & {\color{negred} +2} & {\color{posgreen} +4} & {\color{posgreen} -0.87} & {\color{posgreen} -5} \\
\textbf{Llama} & 39 / 34 & 18.1 / 21.2 & 29 / 36 & 42 / 37 & 10.2 / 11.4 & 29 / 35 \\
\textbf{(70b)} & {\color{negred} -5} & {\color{negred} +3.04} & {\color{negred} +8} & {\color{negred} -4} & {\color{negred} +1.20} & {\color{negred} +6} \\
\textbf{Qwen} & 34 / 30 & 21.6 / 34.8 & 42 / 55 & 40 / 37 & 15.3 / 15.8 & 36 / 40 \\
\textbf{(72b)} & {\color{negred} -4} & {\color{negred} +13.1} & {\color{negred} +13} & {\color{negred} -3} & {\color{negred} +0.50} & {\color{negred} +5} \\
\textbf{GPT} & 53 / 41 & 9.7 / 13.4 & 15 / 26 & 47 / 43 & 7.3 / 7.9 & 24 / 28 \\
\textbf{4.1} & {\color{negred} -12} & {\color{negred} +3.68} & {\color{negred} +11} & {\color{negred} -4} & {\color{negred} +0.66} & {\color{negred} +4} \\
\rowcolor{rowgray} 
& \multicolumn{6}{c}{\textbf{OOD Entity Substitution} (Original / Entity subs)} \\
\rowcolor{headergray} 
& Acc & RMSE & SMAPE & Acc & RMSE & SMAPE\\
\textbf{Gemini} & 52 / 58 & 11.0 / 9.4 & 18 / 13 & 51 / 60 & 5.6 / 3.9 & 23 / 14 \\
\textbf{2.5 Flash}& {\color{posgreen} +7} & {\color{posgreen} -1.61} & {\color{posgreen} -5} & {\color{posgreen} +9} & {\color{posgreen} -1.69} & {\color{posgreen} -9} \\
\textbf{Llama} & 39 / 38 & 18.1 / 18.2 & 29 / 28 & 42 / 41 & 10.2 / 9.3 & 29 / 28 \\
\textbf{(70b)} & {\color{negred} -1} & {\color{negred} +2.60} & {\color{posgreen} -1} & {\color{negred} -0} & {\color{posgreen} -0.92} & {\color{posgreen} -1} \\
\textbf{Qwen} & 34 / 34 & 21.6 / 27.8 & 42 / 42 & 40 / 38 & 15.3 / 17.7 & 36 / 38 \\
\textbf{(72b)} & {\color{negred} 0} & {\color{negred} +6.21} & {\color{negred} 0} & {\color{negred} -2} & {\color{negred} +2.47} & {\color{negred} +2} \\
\textbf{GPT} & 53 / 43 & 9.7 / 9.4 & 15 / 15 & 47 / 47 & 7.3 / 5.7 & 24 / 21 \\
\textbf{4.1} & {\color{negred} -10} & {\color{posgreen} -0.27} & {\color{negred} 0} & {\color{posgreen} +1} & {\color{posgreen} -1.63} & {\color{posgreen} -3} \\
\rowcolor{rowgray} 
& \multicolumn{6}{c}{\textbf{Entity Entanglement} (Original / Entangled)} \\
\rowcolor{headergray} 
& Acc & RMSE & SMAPE & Acc & RMSE & SMAPE\\
\textbf{Gemini} & 52 / 51 & 11.0 / 13.3 & 18 / 21 & 51 / 51 & 5.6 / 5.4 & 23 / 18 \\
\textbf{2.5 Flash}& {\color{negred} -1} & {\color{negred} +2.34} & {\color{negred} +3} & {\color{posgreen} +0} & {\color{posgreen} -0.19} & {\color{posgreen} -5} \\
\textbf{Llama} & 39 / 34 & 18.1 / 22.1 & 29 / 40 & 42 / 40 & 10.2 / 10.8 & 29 / 35 \\
\textbf{(70b)} & {\color{negred} -5} & {\color{negred} +3.99} & {\color{negred} +12} & {\color{negred} -1} & {\color{negred} +0.55} & {\color{negred} +6} \\
\textbf{Qwen} & 34 / 29 & 21.6 / 31.3 & 42 / 60 & 40 / 34 & 15.3 / 17.3 & 36 / 48 \\
\textbf{(72b)} & {\color{negred} -4} & {\color{negred} +9.72} & {\color{negred} +18} & {\color{negred} -5} & {\color{negred} +2.07} & {\color{negred} +12} \\
\textbf{GPT} & 53 / 49 & 9.7 / 11.5 & 15 / 20 & 47 / 44 & 7.3 / 8.0 & 24 / 29 \\
\textbf{4.1} & {\color{negred} -4} & {\color{negred} +1.82} & {\color{negred} +5} & {\color{negred} -3} & {\color{negred} +0.72} & {\color{negred} +5} \\\bottomrule 
\end{tabular}}
\vspace{-1.0em}
\caption{\small Original v/s Perturbed Comparison of Performance for Cricket }
\vspace{-1.0em}
\label{tab:cricket_perturbation}
\end{table}
\begin{table}[t!]
\scriptsize
\setlength{\aboverulesep}{0.25pt}
\setlength{\belowrulesep}{0.25pt}
\setlength{\tabcolsep}{5pt}
\centering
% \resizebox{0.85\linewidth}{!}{%
\begin{tabular}{l|ccc}
\toprule
\rowcolor{headergray}
\textbf{Models} & \textbf{Acc} & \textbf{RMSE} & \textbf{SMAPE} \\
\midrule

\multicolumn{4}{c}{\cellcolor{rowgray}\textbf{Anonymization (Original / Anonymized)}} \\
\addlinespace[1pt]

\textbf{Gemini} & 62 / 35 & 1.59 / 4.41 & 21 / 93 \\
& {\color{negred}-27} & {\color{negred}+2.81} & {\color{negred}+72} \\
\addlinespace[1pt]

\textbf{Llama (70b)} & 30 / 22 & 4.10 / 5.57 & 61 / 87 \\
& {\color{negred}-7} & {\color{negred}+1.47} & {\color{negred}+26} \\
\addlinespace[1pt]

\textbf{Qwen (72b)} & 31 / 22 & 2.65 / 4.84 & 65 / 103 \\
& {\color{negred}-9} & {\color{negred}+2.19} & {\color{negred}+38} \\
\addlinespace[1pt]

\textbf{GPT-4.1} & 47 / 29 & 1.70 / 4.79 & 36 / 99 \\
& {\color{negred}-18} & {\color{negred}+3.10} & {\color{negred}+63} \\
\midrule

\multicolumn{4}{c}{\cellcolor{rowgray}\textbf{OOD Entity Substitution (Original / Entity subs)}} \\
\addlinespace[1pt]

\textbf{Gemini} & 62 / 41 & 1.59 / 3.88 & 21 / 74 \\
& {\color{negred}-21} & {\color{negred}+2.29} & {\color{negred}+53} \\
\addlinespace[1pt]

\textbf{Llama (70b)} & 30 / 27 & 4.10 / 5.55 & 61 / 96 \\
& {\color{negred}-3} & {\color{negred}+1.45} & {\color{negred}+35} \\
\addlinespace[1pt]

\textbf{Qwen (72b)} & 31 / 21 & 2.65 / 4.31 & 65 / 90 \\
& {\color{negred}-11} & {\color{negred}+1.66} & {\color{negred}+26} \\
\addlinespace[1pt]

\textbf{GPT-4.1} & 47 / 29 & 1.70 / 3.98 & 36 / 78 \\
& {\color{negred}-18} & {\color{negred}+2.28} & {\color{negred}+42} \\

\bottomrule
\end{tabular}
% }
\vspace{-1.0em}
\caption{ \small Original v/s Perturbed Comparison of Performance for Basketball}
\label{tab:basketball_perturbations}
\vspace{-2.0em}
\end{table}

\section{Entity-form Robustness}
To further comprehensively understand whether models genuinely \emph{reason} over input rather than keying on brittle surface cues, we probe \emph{form robustness}: sensitivity to changes in phrasing and entity surface forms. Concretely, we introduce 3  perturbation types on the extented version of masked dataset for both cricket and basketball as a baseline. The perturbations introduced are as follows:
\vspace{-0.5em}
% \begin{itemize}
    % \item 

    \paragraph{A. Entity Substitution:} Player names are replaced with \textit{synthetic but plausible, human-sounding names} drawn from entirely different domains: for example, “Arun Hiachak to Lionel Cristiano” instead of “Muzarabani to Tamim.”\footnote{Lionel Cristiano: \textbf{Lionel} Messi + \textbf{Cristiano} Ronaldo, two very famous soccer players.}

    \paragraph{B. Anonymization:} Names of all player mentions are replaced with symbolic placeholder (e.g., \verb|player1|, \verb|player2|). This process mirrors the \textit{delexicalization} strategy used in \cite{suntwal-etal-2019-importance}, to encourage the model to focus on relational and factual reasoning rather than lexical memorization.
    % \p

    \paragraph{C. Entity-Entanglement:} This perturbation is specific to cricket, as basketball commentary does not involve distinct, role-dependent entities. Cricket commentary lines are paraphrased to break the standard \textit{``<bowler> to <batsman>''} pattern by embedding player names within descriptive narrative (as shown in Appendix \ref{appendix:pert}). As a result, player roles are no longer explicitly stated and must be inferred from broader contextual cues. This requires models to reason holistically across multiple commentary lines to correctly assign roles and maintain scorecard consistency.
% \end{itemize}

\subsection{Results and Analysis}
Figure~\ref{fig:overcount-percentage} and Tables~\ref{tab:cricket_perturbation},~\ref{tab:basketball_perturbations} show that perturbations do not simply degrade performance but systematically shift models between hallucination (overcounting)~\cite{10.1145/3571730} and omission (undercounting), depending on the model, sport, role, and perturbation type. To capture this directional behavior, which is not reflected by standard accuracy or numeric error metrics, we introduce the Hallucination--Omission Index (HOI):
{\setlength{\abovedisplayskip}{4pt}
 \setlength{\belowdisplayskip}{4pt}
\[
\mathrm{HOI} = \text{Over-Count\%} - \text{Under-Count\%},
\]}
where Over-Count\% and Under-Count\% denote the proportions of table cells whose predicted values exceed or fall below the ground truth, respectively.

 % Lollipop Plots
% --- DATA DEFINITIONS ---
\pgfplotstableread{
    x  p  Over   Under   OverRMS UnderRMS
    1  0  32.92  15.25   11.63   16.96
    1  1  26.99  14.46   9.60    16.71
    1  2  31.50  17.03   12.01   23.04
    1  3  29.50  17.54   13.09   22.34
    2  0  45.58  20.87   20.87   28.78
    2  1  45.06  21.22   20.94   38.50
    2  2  54.13  16.61   25.89   47.10
    2  3  48.68  21.31   24.49   52.82
    3  0  28.21  33.14   18.55   22.91
    3  1  32.06  29.89   16.90   24.72
    3  2  33.55  32.73   21.53   27.13
    3  3  31.11  35.13   21.20   25.98
    4  0  18.42  28.28   15.26   11.84
    4  1  23.86  32.88   14.95   9.39
    4  2  22.24  28.72   17.76   13.12
    4  3  25.25  33.30   19.31   14.36
}\batsmanData

\pgfplotstableread{
    x  p  Over   Under   OverRMS UnderRMS
    1  0  36.85   3.56    7.13    3.56
    1  1  24.91   3.87    5.29    3.86
    1  2  30.36   4.54    6.21    4.53
    1  3  28.52   5.02    6.20    5.02
    2  0  21.55   38.81   5.06    20.40
    2  1  19.89   41.96   4.39    23.20
    2  2  37.28   28.49   9.30    20.10
    2  3  27.87   35.13   6.55    21.35
    3  0  16.40   42.11   4.35    14.01
    3  1  17.75   40.90   4.64    12.83
    3  2  28.14   31.65   7.10    12.87
    3  3  16.32   46.33   4.74    15.55
    4  0  8.51    44.65   2.53    10.29
    4  1  11.79   40.81   3.64    7.97
    4  2  13.19   43.05   3.81    10.19
    4  3  9.68    47.30   3.21    10.88
}\bowlerData

\pgfplotstableread{
    x  p  Over    Under   OverRMS UnderRMS
    1  0  30.476  7.693   2.726   1.557
    1  1  46.899  9.888   5.945   2.566
    1  3  42.585  19.305  5.307   3.451
    2  0  29.73   38.29   3.24    3.02
    2  1  34.00   39.66   4.68    4.00
    2  3  35.17   41.84   5.28    5.10
    3  0  9.00    61.00   1.85    5.14
    3  1  32.00   39.00   4.75    6.26
    3  3  22.00   54.00   5.43    7.55
    4  0  16.84   36.05   1.90    2.47
    4  1  39.73   25.74   4.96    3.56
    4  3  38.99   30.38   5.74    5.09
}\basketballData

% --- PLOTTING MACRO ---
\newcommand{\plotdumbbells}[3]{
    % 1. Connector Line (Error Bars)
    \pgfplotsinvokeforeach{0,1,2,3}{
        \addplot+[only marks, mark=none, mark size=0pt, forget plot, 
            error bars/.cd, y dir=both, y explicit, 
            error bar style={solid, line width=1pt}, error mark=none] 
            table [x expr={\thisrow{x} + (##1*0.2 - 0.3)}, 
            y expr={\thisrow{p}==##1 ? (\thisrow{#2} + \thisrow{#3})/2 : nan}, 
            y error expr={abs(\thisrow{#2} - \thisrow{#3})/2}] {#1};
    }
    % 2. Over Markers (Filled Circle with Black Border)
    \pgfplotsinvokeforeach{0,1,2,3}{
        \addplot+[only marks, mark=*, mark size=2pt, 
            mark options={draw=black, fill=., line width=0.5pt}] 
        table [x expr={\thisrow{x} + (##1*0.2 - 0.3)}, 
        y expr={\thisrow{p}==##1 ? \thisrow{#2} : nan}] {#1};
    }
    % 3. Under Markers (Open Circle)
    \pgfplotsinvokeforeach{0,1,2,3}{
        \addplot+[forget plot, only marks, mark=o, thick, mark size=2pt] 
        table [x expr={\thisrow{x} + (##1*0.2 - 0.3)}, 
        y expr={\thisrow{p}==##1 ? \thisrow{#3} : nan}] {#1};
    }
}
\begin{figure*}[h!]
    \centering
    \begin{tikzpicture}
        \begin{groupplot}[
            group style={
                group size=3 by 1,
                horizontal sep=0.125cm,
                vertical sep=0cm,
                y descriptions at=edge left,
            },
            width=0.4\linewidth,
            height=5.0cm,
            ymin=0, ymax=70,
            xmin=0.5, xmax=4.5,
            xtick={1,2,3,4},
            xticklabels={Gemini, Qwen, Llama, GPT},
            xticklabel style={
                font=\footnotesize, 
                rotate=10, 
                anchor=north east, 
                inner sep=2pt, 
                yshift=2pt
            },
            yticklabel style={font=\footnotesize},
            ylabel style={
                font=\bfseries\small, 
                at={(axis description cs:-0.09,0.5)}, 
                anchor=south
            },
            ylabel={Percentage},
            title style={font=\bfseries\small, yshift=-1ex},
            ymajorgrids=true,
            grid style={dashed, gray!20},
            axis line style={draw=gray!50},
            cycle list={
                colorOrig\\
                colorCount\\
                colorPara\\
                colorAnon\\
            },
            every axis plot/.append style={thick},
        ]
            % --- Plot 1: Batsman ---
            \nextgroupplot[title={(a) Batsman}]
            \plotdumbbells{\batsmanData}{Over}{Under}
            % --- Plot 2: Bowler ---
            \nextgroupplot[title={(b) Bowler}]
            \plotdumbbells{\bowlerData}{Over}{Under}
            % --- Plot 3: Basketball ---
            \nextgroupplot[title={(c) Basketball}]
            \plotdumbbells{\basketballData}{Over}{Under}
        \end{groupplot}
    \end{tikzpicture}
    % --- Legend ---
    \vspace{-0.1cm}
    \begin{center}
        \footnotesize
        % Row 1: Perturbations
        \textbf{Perturbation:} \enskip
        \textcolor{colorOrig}{\rule[0.2ex]{1.2ex}{1.2ex}} Orig \enskip
        \textcolor{colorCount}{\rule[0.2ex]{1.2ex}{1.2ex}} OOD Entity Subst. \enskip
        \textcolor{colorPara}{\rule[0.2ex]{1.2ex}{1.2ex}} Entity Entanglement \enskip
        \textcolor{colorAnon}{\rule[0.2ex]{1.2ex}{1.2ex}} Anon
        \quad  \quad \quad \quad
        % Row 2: Markers
        \textbf{Marker:} \enskip 
        \mbox{$\bullet$ Over} \quad 
        \mbox{$\circ$ Under}
    \end{center}
    \vspace{-0.9em}
    \caption{Over- \& Under-Count Percentage in cells for Batsman (a), Bowler (b), and Basketball (c).}
    \label{fig:overcount-percentage}
    \vspace{-1em}
\end{figure*}

\paragraph{A. Entity Substitution (Out of Distribution).}
Entity substitution benefits Gemini in cricket but degrades performance in basketball. In cricket, replacing entities with OOD names improves Gemini’s batsman accuracy by \textbf{6pp} and yields an approximate \textbf{15\% reduction in RMSE}, while decreasing batsman HOI by about \textbf{5 percentage pp} (Figure~\ref{fig:overcount-percentage}, Table \ref{tab:cricket_perturbation}). In contrast, the same perturbation in basketball reduces accuracy by \textbf{21pp}, leads to an approximate \textbf{140\% increase in RMSE}, and increases HOI by roughly \textbf{16 pp} (Figure~\ref{fig:overcount-percentage}; Table~\ref{tab:basketball_perturbations}). This divergence suggests that cricket’s summary-driven structure remains stable under name substitution, whereas basketball’s dense, entity-linked updates are more sensitive to disrupted lexical anchors, consistent with prior observations on shortcut reliance \cite{Kandpal2022DeduplicatingTD}.

\paragraph{B. Anonymization.}
Anonymization systematically increases hallucination, with model-specific effects. For Gemini batsmen in cricket, anonymization yields a marginal \textbf{+1pp} gain in accuracy while causing an approximate \textbf{19\% increase in RMSE} and a \textbf{worsening in SMAPE}, indicating more correct cell placements but poorer numerical values (hallucinated precision; Table~\ref{tab:cricket_perturbation}). In contrast, for GPT-4.1 batsmen, anonymization reduces accuracy by \textbf{12pp} and leads to an approximate \textbf{38\% increase in RMSE}, while HOI remains negative with a small shift toward zero (about \textbf{+2pp}), reflecting omission-dominant failures due to withheld updates (Figure~\ref{fig:overcount-percentage}), a conservative strategy documented in prior work \cite{huang2023grounded}. In basketball, anonymization is catastrophic across models, with Gemini exhibiting a \textbf{27pp} accuracy drop alongside an approximate \textbf{180\% increase in RMSE} and a large increase in hallucination (Table~\ref{tab:basketball_perturbations}, Figure \ref{fig:overcount-percentage}), confirming that removing lexical anchors in event-heavy commentary strongly amplifies hallucinated additions.

\paragraph{C. Entity Entanglement.}
 Entity entanglement is the most disruptive perturbation in cricket and frequently flips error direction. For Qwen batsmen, paraphrasing increases hallucination substantially, raising HOI by approximately \textbf{+13pp} and causing an approximate \textbf{24\% increase in overcount RMSE} (Table~\ref{tab:cricket_perturbation}; Figure~\ref{fig:overcount-percentage}). For bowlers, the same perturbation can flip HOI from negative to positive, indicating a transition from omission to hallucination driven by event misattribution rather than token-level errors. Further Table \ref{tab:cricket_extra_missing} shows players being reported under incorrect player rows, demonstrating that breaking lexical role cues disrupts entity grounding and coreference, a known failure mode in contextualized models \cite{Joshi2019BERTFC}.

Joint analysis of HOI with RMSE and SMAPE reveals three recurring model behaviors. \textbf{Hallucination-prone} models (Gemini, often Qwen) show positive HOI and rising numeric error under identity or role disruption, with cases of hallucinated precision (e.g., Gemini batsmen exhibit a \textbf{5pp} HOI decrease under anonymization while RMSE increases). \textbf{Omission-dominant} models (GPT-4.1) maintain negative HOI across perturbations and trade accuracy for conservative numeric estimates. \textbf{Unstable} models (Qwen, LLaMA) exhibit large HOI swings under paraphrase or anonymization, indicating brittle entity--role grounding.

Accuracy alone obscures these differences: GPT-4.1 batsmen can lose \textbf{12pp} accuracy while numeric error improves, whereas Gemini batsmen under anonymization gain \textbf{1pp} accuracy with worse RMSE/SMAPE. These patterns are consistent across cricket and basketball (Tables~\ref{tab:cricket_perturbation}--\ref{tab:basketball_perturbations}; Figure~\ref{fig:overcount-percentage}).

\paragraph{Temporal Robustness.} We also studied \textbf{how performance varies as the input horizon grows} and
whether this temporal effect depends on the presence of summary cues using partial inputs, results of which has been highlighted in Appendix \ref{sec:temporal_robustness}. However, basketball games are kept intact without any partial inputs.

\section{Effect of Memorization}
Since ESPN is a widely scraped web source, we explicitly probe potential data contamination using a chronological split evaluation. For cricket, where commentaries span a long temporal range (2006--2025), we partition samples into pre- and post-2025 sets (January--May 2025), treating the latter as a held-out proxy for post-training evaluation for most models. All models are evaluated with identical prompting across splits. Reported knowledge cutoffs are GPT-4.1 (July 2024), Gemini-2.0 Flash (January 2025), Qwen2.5-72B (September 2024), and LLaMA~3.3-70B (December 2023).

This setup reveals patterns consistent with exposure-driven effects in cricket. GPT-4.1 shows an approximately \textbf{12pp} drop in batsman accuracy on post-2025 data, accompanied by only a small RMSE change, indicating increased withholding of uncertain updates rather than consistent state maintenance. In contrast, Gemini-2.5-Flash improves by approximately \textbf{3pp} in batsman accuracy, with substantial reductions in numerical error (batsman RMSE \textbf{6.06} to \textbf{4.90}, bowler RMSE \textbf{3.90} to \textbf{1.63}). Structural consistency also improves: missing and extra rows drop (9\% to 2\%) on newer data, suggesting improved entity-level state tracking rather than memorized recall. These trends persist under EntityCoT prompting: GPT-4.1 exhibits a large RMSE reduction (9.58 to 6.34) but only modest accuracy gains. Across other baselines, GPT-4.1 consistently shows the smallest accuracy improvement from CoT relative to other models (Table \ref{tab:unified_results}), indicating that memorization effects are orthogonal to prompting strategy. While task decomposition mitigates within-sample multi-entity interference, it does not eliminate biases introduced by pretraining exposure. The absence of older basketball data further supports this interpretation, as basketball performance primarily reflects generalization rather than recall. Together, these findings motivate the use of chronological splits alongside OOD entity perturbations as practical diagnostics for memorization bias in long-context summarization.

\section{Related Work}
\label{sec:related}
\vspace{-0.5em}
A large body of work demonstrates that high average performance can often mask underlying fragilities and reliance on spurious correlations. Methodologies have evolved to include behavioral testing with checklists to evaluate specific linguistic capabilities \cite{ribeiro-etal-2020-beyond}, adversarial and counterfactual examples to assess sensitivity to minor input perturbations \cite{jia-liang-2017-adversarial, mccoy-etal-2019-right, kaushik-etal-2020-learning}, and analyses of long-context reasoning to identify performance degradation over extended inputs \cite{liu-etal-2024-lost}. These diagnostic approaches are essential for understanding model limitations and are increasingly important for information-centric tasks, where reliability and faithfulness to source material are paramount.

The task of text-to-table (T2T) generation has similarly matured from a structured information extraction (IE) problem to a more complex reasoning challenge. Recent approaches have tackled increasingly complex settings where table values must be derived through aggregation, counting, and tracking state changes over long, narrative-style texts. Systems now address nuanced domains like sports analytics \citep{hu2024sportsmetricsblendingtextnumerical} and finance \citep{zhu2024tat}, generate complex hierarchical tables \citep{cheng2021hitab} and employ sophisticated schema-guided pipelines or iterative event extraction \citep{ahuja-etal-2025-map, t3}. As T2T systems take on these more demanding reasoning tasks, the need for robust evaluation, drawing on the principles of model auditing, becomes indispensable for ensuring their reliability.

\section{Conclusion}
\label{sec:conclusion}
We present \textsc{SporTabSet}, a diagnostic benchmark for long-context text-to-table generation, and used it to study how contemporary LLMs summarize evolving sports narratives into structured representations. Across two complementary narrative types, we evaluated a range of modeling strategies, including decomposition-based pipelines, entity-centric generation, and symbolic tuple extraction. 

Our findings lead to three key takeaways. First, task decomposition consistently improves performance, and its benefits arise primarily from mitigating multi-entity interference while enhancing local arithmetic or counting ability across smaller inputs also helpful. Second, models remain brittle under semantics-preserving perturbations, revealing that surface-level cues and memorized patterns play a substantial role in apparent reasoning success. Third, methods that achieve the strongest aggregate performance often incur high costs in reliability and implementation complexity, highlighting a gap between benchmark accuracy and deployable systems. 

Taken together, these results suggest that progress in long-context tabular summarization will require models that can natively maintain structured, entity-consistent state over extended inputs, rather than relying on external memory traces, or brittle surface cues. SporTabSet provides a controlled setting for studying these challenges, and we hope it will inform the development of foundational models and evaluation paradigms that treat multi-entity memory and robustness as first-class objectives in structured generation.

\section*{Limitations}

\label{sec:limitations}

Our study is intentionally diagnostic and scoped to understanding long-context text-to-table generation under controlled, multi-entity settings. While SporTabSet spans two complementary sports domains, both cricket and basketball follow structured narrative conventions and benefit from dense, verifiable supervision. Extending this analysis to other high-impact domains such as finance, healthcare, or scientific reporting remains challenging due to limited public availability of long-form narratives paired with reliable, fine-grained ground truth. Future work that incorporates privacy-preserving data releases or synthetic-but-faithful benchmarks could help bridge this gap.

Our evaluation focuses on stateful numerical tables with fixed schemas and does not consider more complex table structures, such as hierarchical headers, merged cells, temporal indices, or heterogeneous data types (e.g., categorical fields or free-text cells). These structural dimensions introduce additional challenges that are orthogonal to multi-entity state tracking and remain unexplored in this work. Incorporating such table representations would enable a more comprehensive assessment of structured generation capabilities.

Although we explicitly probe memorization effects through chronological splits and entity-level perturbations, we do not attempt to mitigate biases introduced by pretraining on widely scraped sources. Moreover, samples in the old and new splits may differ in inherent difficulty due to stylistic or contextual variation. We partially control for this by maintaining comparable input lengths, schema structure, and event density across splits, but residual confounds may remain. Addressing memorization at the model or data level rather than diagnosing its effects remains an important direction for future research.

Finally, our analysis emphasizes correctness, numerical fidelity, and robustness, but does not directly evaluate downstream utility or human trust in generated tables. Nor do we propose new architectural mechanisms for persistent memory; instead, we focus on characterizing the limitations of existing prompting and decomposition strategies. Future work could translate these findings into architectural or training interventions that enable models to natively maintain structured, entity-consistent state over long contexts, reducing reliance on external prompting scaffolds or symbolic post-processing.

% \section*{Acknowledgement}
% We thank the members of the CoRAL\footnote{https://coral-lab-asu.github.io/}, ASU for their valuable feedback and thoughtful suggestions throughout this project, which materially improved the clarity and rigor of our work. We also thank the Sol Computer cluster at Arizona State University for providing the compute resources that made our experiments possible.

\section*{Ethics Statement}
All data used in this study was collected exclusively from publicly accessible webpages on ESPNcricinfo \footnote{https://www.espncricinfo.com/}, a widely used platform that provides live commentary and statistics on cricket matches. The dataset consists of ball-by-ball textual descriptions and associated player names, which are already publicly available in broadcast and written formats. No private, sensitive, or personally identifiable information beyond publicly known professional athlete names was collected. The scraping process followed a minimal and non-intrusive approach, ensuring that only relevant gameplay commentary was retained while discarding extraneous content such as advertisements, banter, or non-match updates

The resulting dataset was used strictly for academic research purposes, with the goal of creating a benchmark to evaluate language model robustness in text-to-table generation. We recognize that the use of real player names introduces potential concerns regarding memorization or unintended associations. To mitigate this, we explicitly designed robustness experiments that include anonymization and entity substitution, ensuring that the benchmark can be studied without reliance on individual identities. This research does not attempt to commercially exploit the data, nor does it aim to generate derivative works that could misrepresent or harm the reputations of individual players, teams, or organizations. All outputs remain within the scope of scholarly inquiry and reproducible evaluation. The dataset and benchmark are shared under fair-use provisions for research, with appropriate attribution to the original source platform. We have used AI tools such as ChatGPT for assistance in writing, find relevant literature and assistance in coding.

\section*{Licence Statement}
The benchmark artifacts (ground-truth tables, perturbations, and evaluation code) are released under the Creative Commons Attribution–NonCommercial–ShareAlike 4.0 International (CC BY-NC-SA 4.0) license. This permits use, distribution, and adaptation for academic research purposes only, provided that appropriate credit is given and derivative works are shared under the same terms.

The dataset includes structured representations derived from commentary text publicly available on ESPNcricinfo. Original commentary remains under ESPNcricinfo copyright and is redistributed here solely under fair-use provisions for non-commercial research. Commercial use of these artifacts is strictly prohibited.

\bibliography{custom}

\begin{thebibliography}{34}
\providecommand{\natexlab}[1]{#1}

\bibitem[{Ahuja et~al.(2025)Ahuja, Bardoliya, Baral, and Gupta}]{ahuja-etal-2025-map}
Naman Ahuja, Fenil Bardoliya, Chitta Baral, and Vivek Gupta. 2025.
\newblock \href {https://doi.org/10.18653/v1/2025.acl-long.1460} {Map{\&}make: Schema guided text to table generation}.
\newblock In \emph{Proceedings of the 63rd Annual Meeting of the Association for Computational Linguistics (Volume 1: Long Papers)}, pages 30249--30262, Vienna, Austria. Association for Computational Linguistics.

\bibitem[{Ball et~al.(2024)Ball, Chen, and Herley}]{ball2024can}
Thomas Ball, Shuo Chen, and Cormac Herley. 2024.
\newblock Can we count on llms? the fixed-effect fallacy and claims of gpt-4 capabilities.
\newblock \emph{arXiv preprint arXiv:2409.07638}.

\bibitem[{Cheng et~al.(2021)Cheng, Dong, Wang, Jia, Guo, Gao, Han, Lou, and Zhang}]{cheng2021hitab}
Zhoujun Cheng, Haoyu Dong, Zhiruo Wang, Ran Jia, Jiaqi Guo, Yan Gao, Shi Han, Jian-Guang Lou, and Dongmei Zhang. 2021.
\newblock Hitab: A hierarchical table dataset for question answering and natural language generation.
\newblock \emph{arXiv preprint arXiv:2108.06712}.

\bibitem[{Deng et~al.(2024{\natexlab{a}})Deng, Chan, Wang, Sun, Fan, Zheng, Yim, and Song}]{deng-etal-2024-text}
Zheye Deng, Chunkit Chan, Weiqi Wang, Yuxi Sun, Wei Fan, Tianshi Zheng, Yauwai Yim, and Yangqiu Song. 2024{\natexlab{a}}.
\newblock \href {https://doi.org/10.18653/v1/2024.emnlp-main.523} {Text-tuple-table: Towards information integration in text-to-table generation via global tuple extraction}.
\newblock In \emph{Proceedings of the 2024 Conference on Empirical Methods in Natural Language Processing}, pages 9300--9322, Miami, Florida, USA. Association for Computational Linguistics.

\bibitem[{Deng et~al.(2024{\natexlab{b}})Deng, Chan, Wang, Sun, Fan, Zheng, Yim, and Song}]{t3}
Zheye Deng, Chunkit Chan, Weiqi Wang, Yuxi Sun, Wei Fan, Tianshi Zheng, Yauwai Yim, and Yangqiu Song. 2024{\natexlab{b}}.
\newblock \href {https://doi.org/10.18653/v1/2024.emnlp-main.523} {{Text-Tuple-Table: Towards Information Integration in Text-to-Table Generation via Global Tuple Extraction}}.
\newblock In \emph{Proceedings of the 2024 Conference on Empirical Methods in Natural Language Processing}, pages 9300--9322, Miami, Florida, USA. Association for Computational Linguistics.

\bibitem[{Fang et~al.(2024)Fang, Tang, Bi, Qin, Sun, Li, Li, Li, Cong, Lin, Yan, Shi, Song, Liu, and Sun}]{fang2024unimem}
Junjie Fang, Likai Tang, Hongzhe Bi, Yujia Qin, Si~Sun, Zhenyu Li, Haolun Li, Yongjian Li, Xin Cong, Yankai Lin, Yukun Yan, Xiaodong Shi, Sen Song, Zhiyuan Liu, and Maosong Sun. 2024.
\newblock \href {https://openreview.net/forum?id=gQAEGSGVnN} {Unimem: Towards a unified view of long-context large language models}.
\newblock In \emph{First Conference on Language Modeling}.

\bibitem[{Grattafiori et~al.(2024)Grattafiori, Dubey, Jauhri, Pandey, Kadian, Al-Dahle, Letman, Mathur, Schelten, Vaughan, Yang, Fan, Goyal, Hartshorn, Yang, Mitra, Sravankumar, Korenev, Hinsvark, Rao, Zhang, Rodriguez, Gregerson, Spataru, Roziere, Biron, Tang, Chern, Caucheteux, Nayak, Bi, Marra, McConnell, Keller, Touret, Wu, Wong, Ferrer, Nikolaidis, Allonsius, Song, Pintz, Livshits, Wyatt, Esiobu, Choudhary, Mahajan, Garcia-Olano, Perino, Hupkes, Lakomkin, AlBadawy, Lobanova, Dinan, Smith, Radenovic, Guzmán, Zhang, Synnaeve, Lee, Anderson, Thattai, Nail, Mialon, Pang, Cucurell, Nguyen, Korevaar, Xu, Touvron, Zarov, Ibarra, Kloumann, Misra, Evtimov, Zhang, Copet, Lee, Geffert, Vranes, Park, Mahadeokar, Shah, van~der Linde, Billock, Hong, Lee, Fu, Chi, Huang, Liu, Wang, Yu, Bitton, Spisak, Park, Rocca, Johnstun, Saxe, Jia, Alwala, Prasad, Upasani, Plawiak, Li, Heafield, Stone, El-Arini, Iyer, Malik, Chiu, Bhalla, Lakhotia, Rantala-Yeary, van~der Maaten, Chen, Tan, Jenkins, Martin, Madaan, Malo, Blecher,
  Landzaat, de~Oliveira, Muzzi, Pasupuleti, Singh, Paluri, Kardas, Tsimpoukelli, Oldham, Rita, Pavlova, Kambadur, Lewis, Si, Singh, Hassan, Goyal, Torabi, Bashlykov, Bogoychev, Chatterji, Zhang, Duchenne, Çelebi, Alrassy, Zhang, Li, Vasic, Weng, Bhargava, Dubal, Krishnan, Koura, Xu, He, Dong, Srinivasan, Ganapathy, Calderer, Cabral, Stojnic, Raileanu, Maheswari, Girdhar, Patel, Sauvestre, Polidoro, Sumbaly, Taylor, Silva, Hou, Wang, Hosseini, Chennabasappa, Singh, Bell, Kim, Edunov, Nie, Narang, Raparthy, Shen, Wan, Bhosale, Zhang, Vandenhende, Batra, Whitman, Sootla, Collot, Gururangan, Borodinsky, Herman, Fowler, Sheasha, Georgiou, Scialom, Speckbacher, Mihaylov, Xiao, Karn, Goswami, Gupta, Ramanathan, Kerkez, Gonguet, Do, Vogeti, Albiero, Petrovic, Chu, Xiong, Fu, Meers, Martinet, Wang, Wang, Tan, Xia, Xie, Jia, Wang, Goldschlag, Gaur, Babaei, Wen, Song, Zhang, Li, Mao, Coudert, Yan, Chen, Papakipos, Singh, Srivastava, Jain, Kelsey, Shajnfeld, Gangidi, Victoria, Goldstand, Menon, Sharma, Boesenberg,
  Baevski, Feinstein, Kallet, Sangani, Teo, Yunus, Lupu, Alvarado, Caples, Gu, Ho, Poulton, Ryan, Ramchandani, Dong, Franco, Goyal, Saraf, Chowdhury, Gabriel, Bharambe, Eisenman, Yazdan, James, Maurer, Leonhardi, Huang, Loyd, Paola, Paranjape, Liu, Wu, Ni, Hancock, Wasti, Spence, Stojkovic, Gamido, Montalvo, Parker, Burton, Mejia, Liu, Wang, Kim, Zhou, Hu, Chu, Cai, Tindal, Feichtenhofer, Gao, Civin, Beaty, Kreymer, Li, Adkins, Xu, Testuggine, David, Parikh, Liskovich, Foss, Wang, Le, Holland, Dowling, Jamil, Montgomery, Presani, Hahn, Wood, Le, Brinkman, Arcaute, Dunbar, Smothers, Sun, Kreuk, Tian, Kokkinos, Ozgenel, Caggioni, Kanayet, Seide, Florez, Schwarz, Badeer, Swee, Halpern, Herman, Sizov, Guangyi, Zhang, Lakshminarayanan, Inan, Shojanazeri, Zou, Wang, Zha, Habeeb, Rudolph, Suk, Aspegren, Goldman, Zhan, Damlaj, Molybog, Tufanov, Leontiadis, Veliche, Gat, Weissman, Geboski, Kohli, Lam, Asher, Gaya, Marcus, Tang, Chan, Zhen, Reizenstein, Teboul, Zhong, Jin, Yang, Cummings, Carvill, Shepard, McPhie,
  Torres, Ginsburg, Wang, Wu, U, Saxena, Khandelwal, Zand, Matosich, Veeraraghavan, Michelena, Li, Jagadeesh, Huang, Chawla, Huang, Chen, Garg, A, Silva, Bell, Zhang, Guo, Yu, Moshkovich, Wehrstedt, Khabsa, Avalani, Bhatt, Mankus, Hasson, Lennie, Reso, Groshev, Naumov, Lathi, Keneally, Liu, Seltzer, Valko, Restrepo, Patel, Vyatskov, Samvelyan, Clark, Macey, Wang, Hermoso, Metanat, Rastegari, Bansal, Santhanam, Parks, White, Bawa, Singhal, Egebo, Usunier, Mehta, Laptev, Dong, Cheng, Chernoguz, Hart, Salpekar, Kalinli, Kent, Parekh, Saab, Balaji, Rittner, Bontrager, Roux, Dollar, Zvyagina, Ratanchandani, Yuvraj, Liang, Alao, Rodriguez, Ayub, Murthy, Nayani, Mitra, Parthasarathy, Li, Hogan, Battey, Wang, Howes, Rinott, Mehta, Siby, Bondu, Datta, Chugh, Hunt, Dhillon, Sidorov, Pan, Mahajan, Verma, Yamamoto, Ramaswamy, Lindsay, Lindsay, Feng, Lin, Zha, Patil, Shankar, Zhang, Zhang, Wang, Agarwal, Sajuyigbe, Chintala, Max, Chen, Kehoe, Satterfield, Govindaprasad, Gupta, Deng, Cho, Virk, Subramanian, Choudhury,
  Goldman, Remez, Glaser, Best, Koehler, Robinson, Li, Zhang, Matthews, Chou, Shaked, Vontimitta, Ajayi, Montanez, Mohan, Kumar, Mangla, Ionescu, Poenaru, Mihailescu, Ivanov, Li, Wang, Jiang, Bouaziz, Constable, Tang, Wu, Wang, Wu, Gao, Kleinman, Chen, Hu, Jia, Qi, Li, Zhang, Zhang, Adi, Nam, Yu, Wang, Zhao, Hao, Qian, Li, He, Rait, DeVito, Rosnbrick, Wen, Yang, Zhao, and Ma}]{grattafiori2024llama3herdmodels}
Aaron Grattafiori, Abhimanyu Dubey, Abhinav Jauhri, Abhinav Pandey, Abhishek Kadian, Ahmad Al-Dahle, Aiesha Letman, Akhil Mathur, Alan Schelten, Alex Vaughan, Amy Yang, Angela Fan, Anirudh Goyal, Anthony Hartshorn, Aobo Yang, Archi Mitra, Archie Sravankumar, Artem Korenev, Arthur Hinsvark, and 542 others. 2024.
\newblock \href {https://arxiv.org/abs/2407.21783} {The llama 3 herd of models}.
\newblock \emph{Preprint}, arXiv:2407.21783.

\bibitem[{Hu and Wu(2025)}]{hu2025rlstructlightweightreinforcementlearning}
Ruike Hu and Shulei Wu. 2025.
\newblock \href {https://arxiv.org/abs/2512.00319} {Rl-struct: A lightweight reinforcement learning framework for reliable structured output in llms}.
\newblock \emph{Preprint}, arXiv:2512.00319.

\bibitem[{Hu et~al.(2024)Hu, Song, Cho, Wang, Foroosh, Yu, and Liu}]{hu2024sportsmetricsblendingtextnumerical}
Yebowen Hu, Kaiqiang Song, Sangwoo Cho, Xiaoyang Wang, Hassan Foroosh, Dong Yu, and Fei Liu. 2024.
\newblock \href {https://arxiv.org/abs/2402.10979} {Sportsmetrics: Blending text and numerical data to understand information fusion in llms}.
\newblock \emph{Preprint}, arXiv:2402.10979.

\bibitem[{Huang et~al.(2023)Huang, Xia, Shah, Driess, Zeng, Lu, Florence, Mordatch, Levine, Hausman, and brian ichter}]{huang2023grounded}
Wenlong Huang, Fei Xia, Dhruv Shah, Danny Driess, Andy Zeng, Yao Lu, Pete Florence, Igor Mordatch, Sergey Levine, Karol Hausman, and brian ichter. 2023.
\newblock \href {https://openreview.net/forum?id=JCCi58IUsh} {Grounded decoding: Guiding text generation with grounded models for embodied agents}.
\newblock In \emph{Thirty-seventh Conference on Neural Information Processing Systems}.

\bibitem[{Jain et~al.(2024)Jain, Marzoca, and Piccinno}]{jain-etal-2024-structsum}
Parag Jain, Andreea Marzoca, and Francesco Piccinno. 2024.
\newblock \href {https://doi.org/10.18653/v1/2024.acl-long.426} {{STRUCTSUM} generation for faster text comprehension}.
\newblock In \emph{Proceedings of the 62nd Annual Meeting of the Association for Computational Linguistics (Volume 1: Long Papers)}, pages 7876--7896, Bangkok, Thailand. Association for Computational Linguistics.

\bibitem[{Ji et~al.(2023)Ji, Lee, Frieske, Yu, Su, Xu, Ishii, Bang, Madotto, and Fung}]{10.1145/3571730}
Ziwei Ji, Nayeon Lee, Rita Frieske, Tiezheng Yu, Dan Su, Yan Xu, Etsuko Ishii, Ye~Jin Bang, Andrea Madotto, and Pascale Fung. 2023.
\newblock \href {https://doi.org/10.1145/3571730} {Survey of hallucination in natural language generation}.
\newblock \emph{ACM Comput. Surv.}, 55(12).

\bibitem[{Jia and Liang(2017)}]{jia-liang-2017-adversarial}
Robin Jia and Percy Liang. 2017.
\newblock \href {https://doi.org/10.18653/v1/D17-1241} {Adversarial examples for evaluating reading comprehension systems}.
\newblock In \emph{Proceedings of the 2017 Conference on Empirical Methods in Natural Language Processing}, pages 2275--2285, Copenhagen, Denmark. Association for Computational Linguistics.

\bibitem[{Joshi et~al.(2019)Joshi, Levy, Weld, and Zettlemoyer}]{Joshi2019BERTFC}
Mandar Joshi, Omer Levy, Daniel~S. Weld, and Luke Zettlemoyer. 2019.
\newblock \href {https://api.semanticscholar.org/CorpusID:201646551} {Bert for coreference resolution: Baselines and analysis}.
\newblock \emph{ArXiv}, abs/1908.09091.

\bibitem[{Kandpal et~al.(2022)Kandpal, Wallace, and Raffel}]{Kandpal2022DeduplicatingTD}
Nikhil Kandpal, Eric Wallace, and Colin Raffel. 2022.
\newblock \href {https://api.semanticscholar.org/CorpusID:246823128} {Deduplicating training data mitigates privacy risks in language models}.
\newblock \emph{ArXiv}, abs/2202.06539.

\bibitem[{Kaushik et~al.(2020)Kaushik, Hovy, and Lipton}]{kaushik-etal-2020-learning}
Divyansh Kaushik, Eduard Hovy, and Zachary Lipton. 2020.
\newblock \href {https://openreview.net/forum?id=S1x6_2EKDr} {Learning the difference that makes a difference with counterfactually-augmented data}.
\newblock In \emph{International Conference on Learning Representations}.

\bibitem[{Li et~al.(2023)Li, Wang, Shao, Zheng, Wang, and Su}]{li2023sequence}
Tong Li, Zhihao Wang, Liangying Shao, Xuling Zheng, Xiaoli Wang, and Jinsong Su. 2023.
\newblock A sequence-to-sequence\&set model for text-to-table generation.
\newblock In \emph{Findings of the Association for Computational Linguistics: ACL 2023}, pages 5358--5370.

\bibitem[{Li et~al.(2025)Li, Liao, Chen, Wang, and Wang}]{li-etal-2025-dice}
Yiqi Li, Yusheng Liao, Zhe Chen, Yanfeng Wang, and Yu~Wang. 2025.
\newblock \href {https://doi.org/10.18653/v1/2025.emnlp-main.355} {{DICE}: Structured reasoning in {LLM}s through {SLM}-guided chain-of-thought correction}.
\newblock In \emph{Proceedings of the 2025 Conference on Empirical Methods in Natural Language Processing}, pages 6951--6966, Suzhou, China. Association for Computational Linguistics.

\bibitem[{Liu et~al.(2024)Liu, Levonian, Beltagy, Lo, and Smith}]{liu-etal-2024-lost}
Nelson~F. Liu, Kevin Levonian, Iz~Beltagy, Kyle Lo, and Noah~A. Smith. 2024.
\newblock \href {https://doi.org/10.18653/v1/2024.naacl-long.426} {Lost in the middle: How language models use long contexts}.
\newblock In \emph{Proceedings of the 2024 Conference of the North American Chapter of the Association for Computational Linguistics: Human Language Technologies (Volume 1: Long Papers)}, pages 7626--7644, Mexico City, Mexico. Association for Computational Linguistics.

\bibitem[{McCoy et~al.(2019)McCoy, Pavlick, and Linzen}]{mccoy-etal-2019-right}
R.~Thomas McCoy, Ellie Pavlick, and Tal Linzen. 2019.
\newblock \href {https://doi.org/10.18653/v1/N19-1223} {Right for the wrong reasons: Diagnosing syntactic heuristics in natural language inference}.
\newblock In \emph{Proceedings of the 2019 Conference of the North {A}merican Chapter of the Association for Computational Linguistics: Human Language Technologies, Volume 1 (Long and Short Papers)}, pages 2240--2250, Minneapolis, Minnesota. Association for Computational Linguistics.

\bibitem[{OpenAI et~al.(2024)OpenAI, Achiam, Adler, Agarwal, Ahmad, Akkaya, Aleman, Almeida, Altenschmidt, Altman, Anadkat, Avila, Babuschkin, Balaji, Balcom, Baltescu, Bao, Bavarian, Belgum, Bello, Berdine, Bernadett-Shapiro, Berner, Bogdonoff, Boiko, Boyd, Brakman, Brockman, Brooks, Brundage, Button, Cai, Campbell, Cann, Carey, Carlson, Carmichael, Chan, Chang, Chantzis, Chen, Chen, Chen, Chen, Chen, Chess, Cho, Chu, Chung, Cummings, Currier, Dai, Decareaux, Degry, Deutsch, Deville, Dhar, Dohan, Dowling, Dunning, Ecoffet, Eleti, Eloundou, Farhi, Fedus, Felix, Fishman, Forte, Fulford, Gao, Georges, Gibson, Goel, Gogineni, Goh, Gontijo-Lopes, Gordon, Grafstein, Gray, Greene, Gross, Gu, Guo, Hallacy, Han, Harris, He, Heaton, Heidecke, Hesse, Hickey, Hickey, Hoeschele, Houghton, Hsu, Hu, Hu, Huizinga, Jain, Jain, Jang, Jiang, Jiang, Jin, Jin, Jomoto, Jonn, Jun, Kaftan, Łukasz Kaiser, Kamali, Kanitscheider, Keskar, Khan, Kilpatrick, Kim, Kim, Kim, Kirchner, Kiros, Knight, Kokotajlo, Łukasz Kondraciuk,
  Kondrich, Konstantinidis, Kosic, Krueger, Kuo, Lampe, Lan, Lee, Leike, Leung, Levy, Li, Lim, Lin, Lin, Litwin, Lopez, Lowe, Lue, Makanju, Malfacini, Manning, Markov, Markovski, Martin, Mayer, Mayne, McGrew, McKinney, McLeavey, McMillan, McNeil, Medina, Mehta, Menick, Metz, Mishchenko, Mishkin, Monaco, Morikawa, Mossing, Mu, Murati, Murk, Mély, Nair, Nakano, Nayak, Neelakantan, Ngo, Noh, Ouyang, O'Keefe, Pachocki, Paino, Palermo, Pantuliano, Parascandolo, Parish, Parparita, Passos, Pavlov, Peng, Perelman, de~Avila Belbute~Peres, Petrov, de~Oliveira~Pinto, Michael, Pokorny, Pokrass, Pong, Powell, Power, Power, Proehl, Puri, Radford, Rae, Ramesh, Raymond, Real, Rimbach, Ross, Rotsted, Roussez, Ryder, Saltarelli, Sanders, Santurkar, Sastry, Schmidt, Schnurr, Schulman, Selsam, Sheppard, Sherbakov, Shieh, Shoker, Shyam, Sidor, Sigler, Simens, Sitkin, Slama, Sohl, Sokolowsky, Song, Staudacher, Such, Summers, Sutskever, Tang, Tezak, Thompson, Tillet, Tootoonchian, Tseng, Tuggle, Turley, Tworek, Uribe, Vallone,
  Vijayvergiya, Voss, Wainwright, Wang, Wang, Wang, Ward, Wei, Weinmann, Welihinda, Welinder, Weng, Weng, Wiethoff, Willner, Winter, Wolrich, Wong, Workman, Wu, Wu, Wu, Xiao, Xu, Yoo, Yu, Yuan, Zaremba, Zellers, Zhang, Zhang, Zhao, Zheng, Zhuang, Zhuk, and Zoph}]{openai2024gpt4technicalreport}
OpenAI, Josh Achiam, Steven Adler, Sandhini Agarwal, Lama Ahmad, Ilge Akkaya, Florencia~Leoni Aleman, Diogo Almeida, Janko Altenschmidt, Sam Altman, Shyamal Anadkat, Red Avila, Igor Babuschkin, Suchir Balaji, Valerie Balcom, Paul Baltescu, Haiming Bao, Mohammad Bavarian, Jeff Belgum, and 262 others. 2024.
\newblock \href {https://arxiv.org/abs/2303.08774} {Gpt-4 technical report}.
\newblock \emph{Preprint}, arXiv:2303.08774.

\bibitem[{Pietruszka et~al.(2024)Pietruszka, Turski, Borchmann, Dwojak, Nowakowska, Szyndler, Jurkiewicz, and Garncarek}]{pietruszka2024stable}
Micha{\l} Pietruszka, Micha{\l} Turski, {\L}ukasz Borchmann, Tomasz Dwojak, Gabriela Nowakowska, Karolina Szyndler, Dawid Jurkiewicz, and {\L}ukasz Garncarek. 2024.
\newblock Stable: Table generation framework for encoder-decoder models.
\newblock In \emph{Proceedings of the 18th Conference of the European Chapter of the Association for Computational Linguistics (Volume 1: Long Papers)}, pages 2454--2472.

\bibitem[{Qwen et~al.(2025)Qwen, :, Yang, Yang, Zhang, Hui, Zheng, Yu, Li, Liu, Huang, Wei, Lin, Yang, Tu, Zhang, Yang, Yang, Zhou, Lin, Dang, Lu, Bao, Yang, Yu, Li, Xue, Zhang, Zhu, Men, Lin, Li, Tang, Xia, Ren, Ren, Fan, Su, Zhang, Wan, Liu, Cui, Zhang, and Qiu}]{qwen2025qwen25technicalreport}
Qwen, :, An~Yang, Baosong Yang, Beichen Zhang, Binyuan Hui, Bo~Zheng, Bowen Yu, Chengyuan Li, Dayiheng Liu, Fei Huang, Haoran Wei, Huan Lin, Jian Yang, Jianhong Tu, Jianwei Zhang, Jianxin Yang, Jiaxi Yang, Jingren Zhou, and 25 others. 2025.
\newblock \href {https://arxiv.org/abs/2412.15115} {Qwen2.5 technical report}.
\newblock \emph{Preprint}, arXiv:2412.15115.

\bibitem[{Ribeiro et~al.(2020)Ribeiro, Wu, Guestrin, and Singh}]{ribeiro-etal-2020-beyond}
Marco~Tulio Ribeiro, Tongshuang Wu, Carlos Guestrin, and Sameer Singh. 2020.
\newblock \href {https://doi.org/10.18653/v1/2020.acl-main.442} {{C}heck{L}ist: Beyond accuracy in {NLU} evaluation}.
\newblock In \emph{Proceedings of the 58th Annual Meeting of the Association for Computational Linguistics}, pages 4902--4912, Online. Association for Computational Linguistics.

\bibitem[{Schick et~al.(2023)Schick, Dwivedi-Yu, Dess{\`i} et~al.}]{schick2023toolformer}
Timo Schick, Jane Dwivedi-Yu, Roberto Dess{\`i}, and 1 others. 2023.
\newblock Toolformer: Language models can teach themselves to use tools.
\newblock In \emph{Proceedings of the International Conference on Learning Representations (ICLR)}.

\bibitem[{Sundar et~al.(2024)Sundar, Richardson, and Heck}]{sundar2024gtbls}
Anirudh Sundar, Christopher Richardson, and Larry Heck. 2024.
\newblock gtbls: Generating tables from text by conditional question answering.
\newblock \emph{arXiv preprint arXiv:2403.14457}.

\bibitem[{Suntwal et~al.(2019)Suntwal, Paul, Sharp, and Surdeanu}]{suntwal-etal-2019-importance}
Sandeep Suntwal, Mithun Paul, Rebecca Sharp, and Mihai Surdeanu. 2019.
\newblock \href {https://doi.org/10.18653/v1/D19-1340} {On the importance of delexicalization for fact verification}.
\newblock In \emph{Proceedings of the 2019 Conference on Empirical Methods in Natural Language Processing and the 9th International Joint Conference on Natural Language Processing (EMNLP-IJCNLP)}, pages 3413--3418, Hong Kong, China. Association for Computational Linguistics.

\bibitem[{Wang et~al.(2025{\natexlab{a}})Wang, Shen, Mishra, Xu, Teng, and Ding}]{wang2025slotstructuringoutputlarge}
Darren Yow-Bang Wang, Zhengyuan Shen, Soumya~Smruti Mishra, Zhichao Xu, Yifei Teng, and Haibo Ding. 2025{\natexlab{a}}.
\newblock \href {https://arxiv.org/abs/2505.04016} {Slot: Structuring the output of large language models}.
\newblock \emph{Preprint}, arXiv:2505.04016.

\bibitem[{Wang et~al.(2025{\natexlab{b}})Wang, Fu, Cao, Wang, Tian, and Ding}]{wang2025recursivelysummarizingenableslongterm}
Qingyue Wang, Yanhe Fu, Yanan Cao, Shuai Wang, Zhiliang Tian, and Liang Ding. 2025{\natexlab{b}}.
\newblock \href {https://arxiv.org/abs/2308.15022} {Recursively summarizing enables long-term dialogue memory in large language models}.
\newblock \emph{Preprint}, arXiv:2308.15022.

\bibitem[{Wang et~al.(2023)Wang, Dong, Cheng, Liu, Yan, Gao, and Wei}]{wang2023augmenting}
Weizhi Wang, Li~Dong, Hao Cheng, Xiaodong Liu, Xifeng Yan, Jianfeng Gao, and Furu Wei. 2023.
\newblock \href {https://openreview.net/forum?id=BryMFPQ4L6} {Augmenting language models with long-term memory}.
\newblock In \emph{Thirty-seventh Conference on Neural Information Processing Systems}.

\bibitem[{Wei et~al.(2022)Wei, Wang, Schuurmans, Bosma, brian ichter, Xia, Chi, Le, and Zhou}]{cot}
Jason Wei, Xuezhi Wang, Dale Schuurmans, Maarten Bosma, brian ichter, Fei Xia, Ed~H. Chi, Quoc~V Le, and Denny Zhou. 2022.
\newblock \href {https://openreview.net/forum?id=_VjQlMeSB_J} {{Chain of Thought Prompting Elicits Reasoning in Large Language Models}}.
\newblock In \emph{Advances in Neural Information Processing Systems}.

\bibitem[{Wu et~al.(2022)Wu, Zhang, and Li}]{text-2-table-2022-acl}
Xueqing Wu, Jiacheng Zhang, and Hang Li. 2022.
\newblock \href {https://doi.org/10.18653/v1/2022.acl-long.180} {{Text-to-Table: A New Way of Information Extraction}}.
\newblock In \emph{Proceedings of the 60th Annual Meeting of the Association for Computational Linguistics (Volume 1: Long Papers)}, pages 2518--2533, Dublin, Ireland. Association for Computational Linguistics.

\bibitem[{Yao et~al.(2022)Yao, Zhao, Yu, Shafran, Narasimhan, and Cao}]{yao2022react}
Shunyu Yao, Jeffrey Zhao, Dian Yu, Izhak Shafran, Karthik~R Narasimhan, and Yuan Cao. 2022.
\newblock \href {https://openreview.net/forum?id=tvI4u1ylcqs} {React: Synergizing reasoning and acting in language models}.
\newblock In \emph{NeurIPS 2022 Foundation Models for Decision Making Workshop}.

\bibitem[{Zhu et~al.(2024)Zhu, Liu, Feng, Wang, Li, and Chua}]{zhu2024tat}
Fengbin Zhu, Ziyang Liu, Fuli Feng, Chao Wang, Moxin Li, and Tat~Seng Chua. 2024.
\newblock Tat-llm: A specialized language model for discrete reasoning over financial tabular and textual data.
\newblock In \emph{Proceedings of the 5th ACM International Conference on AI in Finance}, pages 310--318.

\end{thebibliography}
\appendix
\label{sec:appendix}

\section{Data Perturbations}
\label{appendix:pert}
\begin{table}[H]
\centering
\small
\begin{tabular}{>{\raggedright\arraybackslash}
p{0.19\columnwidth}
>{\raggedright\arraybackslash}p{0.19\columnwidth}
>{\raggedright\arraybackslash}p{0.19\columnwidth}
>{\raggedright\arraybackslash}p{0.19\columnwidth}}\toprule
\rowcolor[HTML]{C0C0C0}
Original & OOD Entity\newline Sub. & Anon. & Entangle\newline-ment \\\midrule
Muzarabani to Tamim, no run fullish ball on middle and leg, clipped to square leg. 
& \hl{Vinicius Alcaraz} to \hl{Matt Ingebrigtsen}, no run fullish ball on middle and leg, clipped to square leg. 
& \hl{Player3} to \hl{Player7}, no run fullish ball on middle and leg, clipped to square leg. 
& Spotlight swings toward \hl{Muzarabani / Tamim}; no run fullish ball on middle and leg, clipped to square leg. \\

Muzarabani to Tamim, no run full and straighter on off this time. Driven down the ground with the full face of the bat. 
& \hl{Vinicius Alcaraz} to \hl{Matt Ingebrigtsen}, no run full and straighter on off this time. Driven down the ground with the full face of the bat. 
& \hl{Player3} to \hl{Player7}, no run full and straighter on off this time. Driven down the ground with the full face of the bat. 
& Action with \hl{Tamim and Muzarabani}: no run full and straighter on off this time. Driven down the ground with the full face of the bat. \\

Muzarabani to Tamim, no run past the outside edge to the batter. Fullish length in the off-stump channel, as Tamim presents his bat playing in the line and the ball shoots off the pitch. 
& \hl{Vinicius Alcaraz} to \hl{Matt Ingebrigtsen}, no run past the outside edge to the batter. Fullish length in the off-stump channel, as \hl{Matt Ingebrigtsen} presents his bat playing in the line and the ball shoots off the pitch. 
& \hl{Player3} to \hl{Player7}, no run past the outside edge to the batter. Fullish length in the off-stump channel, as \hl{Player7} presents his bat playing in the line and the ball shoots off the pitch. 
& Between \hl{Tamim and Muzarabani}: no run past the outside edge to the batter. Fullish length in the off-stump channel, as \hl{Tamim} presents his bat playing in the line and the ball shoots off the pitch. \\ \bottomrule
\end{tabular}
\caption{Comparison of cricket commentary variations across different transformation types. Highlighted text indicates specific differences from the original commentary in terms of entity substitution (player names), structural changes, and semantic modifications.}
\label{tab:commentary-variations}
\end{table}
We induce three perturbations to study the robustness of the models. \textbf{Anonymizaion} where the named entities for bowler and batsman are replaced by generic placeholders like "\textit{Player1}", "\textit{Player2}" and so on. \textbf{Out of Distribution Entity Substitution} that substituted the entities with made up names generated using amalgamation of different names from other sports other than cricket. \textbf{Entity Entanglement}, where the roles of entities are deliberately obscured to challenge role identification. In the original cricket commentary, predictable structural patterns clearly distinguish batsmen from bowlers (e.g., "Sharma bowls to Kohli" vs "Kohli faces Sharma"). This perturbation disrupts these cues by using ambiguous phrasing and reversed sentence structures, requiring models to maintain holistic contextual understanding rather than relying on positional heuristics. Examples contrasting the original, anonymized, OOD entity-substituted, and entity-entangled versions are shown in Table \ref{tab:commentary-variations} for cricket, while the equivalent perturbations are shown for basketball in Table \ref{tab:commentary-variations-basketball}. Note that this perturbation was applied only to cricket commentary, as basketball play-by-play lacks the same structured role distinctions between offensive and defensive players.
\begin{table}[H]
\centering
\small
\begin{tabular}{>{\raggedright\arraybackslash}
p{0.19\columnwidth}
>{\raggedright\arraybackslash}p{0.19\columnwidth}
>{\raggedright\arraybackslash}p{0.19\columnwidth}
>{\raggedright\arraybackslash}p{0.19\columnwidth}}\toprule
\rowcolor[HTML]{C0C0C0}
Original & OOD Entity\newline Sub. & Anon. \\\midrule
Nic Claxton vs. Brook Lopez (Giannis Antetokounmpo gains possession)
& \hl{Madison Jackson} vs. \hl{Dustin Bolt} (\hl{Rafael Jefferson} gains possession)
& \hl{Player35} vs. \hl{Player4} (\hl{Player9} gains possession)
\\ \midrule
Giannis Antetokounmpo misses 11-foot hook shot
& \hl{Rafael Jefferson} misses 11-foot hook shot
& \hl{Player9} misses 11-foot hook shot
\\ \midrule
James Harden blocks Giannis Antetokounmpo's two point shot
& \hl{Auston Sanders} blocks \hl{Rafael Jefferson's} two point shot
& \hl{Player 14} blocks \hl{Player 9's} two point shot
\\ \bottomrule
\end{tabular}
\caption{Comparison of basketball commentary variations across different transformation types. Highlighted text indicates specific differences from the original commentary in terms of entity substitution (player names), structural changes, and semantic modifications.}
\label{tab:commentary-variations-basketball}
\end{table}
\subsection{Effect on Memorization}
\begin{table}[H]
\centering
\footnotesize 
\setlength{\tabcolsep}{1.5pt} % Tight spacing to prevent resizebox shrinking
\resizebox{\linewidth}{!}{%
\begin{tabular}{l|ccc|ccc}\toprule
\rowcolor[HTML]{EFEFEF} & \multicolumn{3}{c|}{\textbf{Batsman}} & \multicolumn{3}{c}{\textbf{Bowler}} \\ \midrule
\rowcolor[HTML]{EFEFEF} & \multicolumn{6}{c}{\cellcolor[HTML]{EFEFEF}Comparison (Old 2025 / New 2025)}\\
\rowcolor[HTML]{C0C0C0} 
& Acc & RMSE & SMAPE & Acc & RMSE & SMAPE\\

% --- Gemini ---
\textbf{Gemini} & 57 / 60 & 9.1 / 5.8 & 16 / 7 & 60 / 65 & 3.4 / 1.9 & 17 / 6 \\
 \textbf{2.5 Flash} & {\color[HTML]{32CB00} +3} & {\color[HTML]{32CB00} -3.24} & {\color[HTML]{32CB00} -9} & {\color[HTML]{32CB00} +5} & {\color[HTML]{32CB00} -1.51} & {\color[HTML]{32CB00} -11} \\

% --- Llama (70b) ---
\textbf{Llama} & 37 / 36 & 20.3 / 20.2 & 37 / 33 & 36 / 45 & 8.6 / 11.0 & 38 / 25 \\
\textbf{(70b)} & {\color[HTML]{FD6864} -1} & {\color[HTML]{32CB00} -0.14} & {\color[HTML]{32CB00} -4} & {\color[HTML]{32CB00} +9} & {\color[HTML]{FD6864} +2.48} & {\color[HTML]{32CB00} -13} \\

% --- Qwen (72b) ---
\textbf{Qwen} & 31 / 32 & 20.5 / 19.8 & 50 / 42 & 38 / 44 & 8.2 / 11.7 & 39 / 27 \\
\textbf{(72b)} & {\color[HTML]{32CB00} +1} & {\color[HTML]{32CB00} -0.75} & {\color[HTML]{32CB00} -8} & {\color[HTML]{32CB00} +6} & {\color[HTML]{FD6864} +3.57} & {\color[HTML]{32CB00} -12} \\

% --- GPT 4.1 ---
\textbf{GPT} & 52 / 40 & 13.5 / 12.7 & 26 / 20 & 43 / 49 & 8.7 / 6.9 & 37 / 18 \\
\textbf{4.1} & {\color[HTML]{FD6864} -12} & {\color[HTML]{32CB00} -0.81} & {\color[HTML]{32CB00} -6} & {\color[HTML]{32CB00} +6} & {\color[HTML]{32CB00} -1.80} & {\color[HTML]{32CB00} -19} \\\bottomrule
\end{tabular}}
\caption{Comparison of Model performance on Pre-2025 and Post 2025 samples}
\label{tab:old_new}
\end{table}
As presented in Table \ref{tab:old_new}, most models show gains in performance on the new samples. Notably, GPT-4.1 shows a steep drop in accuracy for Batsman tables, signifying the intertwining of reasoning with pre-trained bias for specific entities.
\subsection{Masked and Unmasked Commentary}
\begin{table}[H]
\centering
\footnotesize 
\setlength{\tabcolsep}{1.5pt} % Tight spacing to prevent resizebox shrinking
\resizebox{\linewidth}{!}{%
\begin{tabular}{l|ccc|ccc}\toprule
\rowcolor[HTML]{EFEFEF} & \multicolumn{3}{c|}{\textbf{Batsman}} & \multicolumn{3}{c}{\textbf{Bowler}} \\ \midrule
\rowcolor[HTML]{EFEFEF} & \multicolumn{6}{c}{\cellcolor[HTML]{EFEFEF}Comparison (Masked / Unmasked)}\\
\rowcolor[HTML]{C0C0C0} 
& Acc & RMSE & SMAPE & Acc & RMSE & SMAPE\\

% --- Gemini ---
\textbf{Gemini} & 49 / 86 & 13.82 / 4.26 & 22 / 4 & 43 / 55 & 7.83 / 4.93 & 28 / 15 \\
\textbf{2.5 Flash} & {\color[HTML]{32CB00} +37} & {\color[HTML]{32CB00} -9.56} & {\color[HTML]{32CB00} -18} & {\color[HTML]{32CB00} +12} & {\color[HTML]{32CB00} -2.90} & {\color[HTML]{32CB00} -13} \\

% --- Qwen (72b) ---
\textbf{Qwen} & 34 / 80 & 19.19 / 11.10 & 40 / 12 & 38 / 37 & 12.23 / 23.27 & 35 / 39 \\
\textbf{(72b)} & {\color[HTML]{32CB00} +46} & {\color[HTML]{32CB00} -8.09} & {\color[HTML]{32CB00} -28} & {\color[HTML]{FD6864} -1} & {\color[HTML]{FD6864} +11.04} & {\color[HTML]{FD6864} +4} \\

% --- Llama (70b) ---
\textbf{Llama} & 39 / 85 & 19.76 / 6.71 & 32 / 7 & 40 / 42 & 11.95 / 18.14 & 29 / 30 \\
\textbf{(70b)} & {\color[HTML]{32CB00} +46} & {\color[HTML]{32CB00} -13.05} & {\color[HTML]{32CB00} -25} & {\color[HTML]{32CB00} +2} & {\color[HTML]{FD6864} +6.19} & {\color[HTML]{FD6864} +1} \\

% --- GPT 4.1 ---
\textbf{GPT} & 61 / 89 & 9.82 / 4.68 & 14 / 4 & 51 / 52 & 7.46 / 7.28 & 23 / 22 \\
\textbf{4.1} & {\color[HTML]{32CB00} +28} & {\color[HTML]{32CB00} -5.14} & {\color[HTML]{32CB00} -10} & {\color[HTML]{32CB00} +1} & {\color[HTML]{32CB00} -0.18} & {\color[HTML]{32CB00} -1} \\

\bottomrule
\end{tabular}}
\caption{Analysis of Model performance on Masked and Unmasked Cricket Commentary}
\label{tab:Masked_vs_unmasked}
\end{table}
Table \ref{tab:Masked_vs_unmasked} presents the comparison between masked (removing batsman summary cues)and unmasked commentories. All models shows substantial improvement in performance, validating that summary cues reduce this task of reasoning to simple extraction. Notably, for Gemini, performance on bowler tables also improves, with a significant reduction of error rates.

\section{Error Analysis}

\subsubsection{Error Analysis in Entity Entanglement}
Under the \textbf{Entity-Entangled} condition, models do not make random entity-level errors; instead, they systematically confuse \emph{roles}. Table~\ref{tab:cricket_extra_missing} reports the number of extra rows, missing rows, and cross-role entity overlap between the batsman and bowler scorecards for different models and perturbations in the cricket domain.

We quantify this phenomenon using entity entanglement metrics. \textbf{Batsman Ent\%} measures the proportion of entities in the predicted batsman scorecard that appear in the ground-truth bowler scorecard, while \textbf{Bowler Ent\%} measures the proportion of entities in the predicted bowler scorecard that appear in the ground-truth batsman scorecard. Both metrics are computed using the Jaccard overlap,
$\mathrm{IoU}(A,B) = \frac{|A \cap B|}{|A \cup B|}$.

Across all models, we observe consistently high cross-role overlap under entity-entangled perturbations, indicating a failure to preserve role boundaries. We illustrate this behavior with a concrete example below in Table \ref{tab: batsman_gt_bowler_gt} and \ref{tab: bowler_pred_batsman_pred}

\begin{table}[h!]
\centering
% \resizebox{\linewidth}{!}
{
\begin{tabular}{ll}
\hline
\textbf{Batsman (GT)} & \textbf{Bowler (GT)} \\
\hline
Bagai            & Benn      \\
Bhatti           & Bernard   \\
Cheema           & Rampaul   \\
Chohan           & Sammy     \\
Dhaniram         &           \\
Limbada          &           \\
Patel            &           \\
Surkari          &           \\
TC Bastiampillai &           \\
\hline
\end{tabular}
}
\caption{Ground-truth entities with strictly disjoint batting and bowling roles.}
\label{tab: batsman_gt_bowler_gt}
\end{table}

\paragraph{Model Predictions under Entity Entanglement}

\begin{table}[h!]
\centering
\resizebox{\linewidth}{!}{
\begin{tabular}{ll}
\hline
\textbf{Bowler Pred (Entangled)} & \textbf{Batsman Pred (Entangled)} \\
\hline
Rampaul           & TC Bastiampillai \\
TC Bastiampillai & Rampaul          \\
Cheema           & Cheema           \\
Bernard          & Bernard          \\
Surkari          & Surkari          \\
Limbada          & Bagai            \\
Benn             & Limbada          \\
Sammy            & Benn             \\
Patel            & Sammy            \\
Bhatti           & Dhaniram         \\
Dhaniram         & Patel            \\
Chohan           & Bhatti           \\
Chohan           &                  \\
\hline
\end{tabular}
}
\caption{Predicted scorecards under entity entanglement, showing cross-role mixing.}
\label{tab: bowler_pred_batsman_pred}
\end{table}

\paragraph{Analysis}

In the ground truth, batsmen and bowlers form two strictly disjoint entity sets. Under entity entanglement, however, the model fails to maintain these role distinctions and assigns entities interchangeably across roles.

Specifically, the predicted bowler scorecard contains eight entities that belong to the ground-truth batsman set
(\emph{Bhatti, Cheema, Chohan, Dhaniram, Limbada, Patel, Surkari, TC Bastiampillai}),
yielding a \textbf{61.5\%} Jaccard overlap with the batsman ground truth.

Conversely, the predicted batsman scorecard includes four ground-truth bowlers
(\emph{Benn, Bernard, Rampaul, Sammy}),
resulting in a \textbf{30.7\%} overlap with the bowler ground truth.

This example demonstrates that entity entanglement is fundamentally a \emph{role-confusion failure mode}: the model does not merely misidentify entities, but systematically swaps their semantic roles when contextual signals are entangled.

\subsubsection{Missing vs Extra Rows}
Tables~\ref{tab:cricket_extra_missing} and~\ref{tab:nba} summarize the number of missing and spurious entities produced by different models in the generated summary tables. In the original cricket setting, Gemini exhibits the highest number of missing entities relative to other models, but its performance improves when perturbed commentaries are provided. In contrast, the remaining models remain relatively stable in the cricket domain, showing limited variation in missing or extra entities under perturbations. The missing and extra rows percentages were checked against a range of name-matching thresholds from 0.5 to 0.1 making sure that the missing and extra entities are not comprised of entities that failed the name-match test.

The observed instability under the entity-entangled condition can be attributed to confusion between batsmen and bowlers, where entities are systematically interchanged across roles. In the basketball domain, which contains substantially more entities than cricket, all models exhibit a higher number of missing entities under perturbed settings. Models consistently miss more entities in basketball than in cricket, suggesting that an increase in the number of entities adversely impacts models’ ability to reliably track and preserve entity representations.

\begin{table}[h!]
\centering
\renewcommand{\arraystretch}{1.2}
\setlength{\aboverulesep}{0pt}
\setlength{\belowrulesep}{0.2pt}
\setlength{\tabcolsep}{1.5pt}
\footnotesize
% \resizebox{\linewidth}{!}{%
\begin{tabular}{l|ccc|ccc}\toprule
\rowcolor{rowgray} & \multicolumn{3}{c|}{\textbf{Batsman}} & \multicolumn{3}{c}{\textbf{Bowler}} \\ \midrule

% --- SECTION 1: COUNTERFACTUAL ---
\rowcolor{rowgray} & \multicolumn{6}{c}{\textbf{Entity Subs. (OOD)} (Original / Perturbed)}\\
\rowcolor{headergray} 
\textbf{Model} & Extra\% & Miss\% & Ent\% & Extra\% & Miss\% & Ent\% \\
\textbf{Gemini 2.5} & 0 / 0 & 8 / 1 & 0 / 0 & 0 / 0 & 5 / 0 & 0 / 0 \\
\textit{Delta} & {\color{negred} +0} & {\color{posgreen} -7} & {\color{negred} +0} & {\color{negred} +0} & {\color{posgreen} -5} & {\color{negred} +0} \\
\textbf{Qwen 72b} & 0 / 1 & 3 / 3 & 0 / 0 & 0 / 0 & 2 / 1 & 0 / 0 \\
\textit{Delta} & {\color{negred} +1} & {\color{negred} +0} & {\color{negred} +0} & {\color{negred} +0} & {\color{posgreen} -1} & {\color{negred} +0} \\
\textbf{Llama 70b} & 0 / 0 & 1 / 1 & 0 / 0 & 0 / 0 & 0 / 0 & 0 / 0 \\
\textit{Delta} & {\color{negred} +0} & {\color{negred} +0} & {\color{negred} +0} & {\color{posgreen} -0} & {\color{negred} +0} & {\color{negred} +0} \\
\textbf{GPT 4.1} & 1 / 1 & 1 / 1 & 0 / 0 & 0 / 0 & 0 / 0 & 0 / 0 \\
\textit{Delta} & {\color{negred} +0} & {\color{negred} +0} & {\color{negred} +0} & {\color{negred} +0} & {\color{posgreen} -0} & {\color{negred} +0} \\

% --- SECTION 2: PARAPHRASE ---
\rowcolor{rowgray} 
& \multicolumn{6}{c}{\textbf{Entity Entangled} (Original / Perturbed)} \\
\rowcolor{headergray} 
\textbf{Model} & Extra\% & Miss\% & Ent\% & Extra\% & Miss\% & Ent\% \\
\textbf{Gemini 2.5} & 0 / 6 & 8 / 1 & 0 / 6 & 0 / 6 & 5 / 1 & 0 / 5 \\
\textit{Delta} & {\color{negred} +6} & {\color{posgreen} -7} & {\color{negred} +6} & {\color{negred} +6} & {\color{posgreen} -4} & {\color{negred} +5} \\
\textbf{Qwen 72b} & 0 / 12 & 3 / 8 & 0 / 10 & 0 / 7 & 2 / 10 & 0 / 6 \\
\textit{Delta} & {\color{negred} +12} & {\color{posgreen} +5} & {\color{negred} +10} & {\color{negred} +7} & {\color{negred} +8} & {\color{negred} +6} \\
\textbf{Llama 70b} & 0 / 6 & 1 / 3 & 0 / 5 & 0 / 5 & 0 / 1 & 0 / 4 \\
\textit{Delta} & {\color{negred} +6} & {\color{negred} +2} & {\color{negred} +5} & {\color{negred} +5} & {\color{negred} +1} & {\color{negred} +4} \\
\textbf{GPT 4.1} & 1 / 6 & 1 / 1 & 0 / 4 & 0 / 3 & 0 / 1 & 0 / 2 \\
\textit{Delta} & {\color{negred} +5} & {\color{negred} +0} & {\color{negred} +4} & {\color{negred} +3} & {\color{negred} +1} & {\color{negred} +2} \\

% --- SECTION 3: SYMBOLIC ---
\rowcolor{rowgray} 
& \multicolumn{6}{c}{\textbf{Anonymized} (Original / Perturbed)} \\
\rowcolor{headergray}
\textbf{Model} & Extra\% & Miss\% & Ent\% & Extra\% & Miss\% & Ent\% \\
\textbf{Gemini 2.5} & 0 / 1 & 8 / 2 & 0 / 0 & 0 / 0 & 5 / 1 & 0 / 0 \\ \textit{Delta} & {\color{negred} +1} & {\color{posgreen} -6} & {\color{negred} +0} & {\color{negred} +0} & {\color{posgreen} -4} & {\color{negred} +0} \\
\textbf{Qwen 72b} & 0 / 1 & 3 / 6 & 0 / 0 & 0 / 0 & 2 / 3 & 0 / 0 \\
\textit{Delta} & {\color{posgreen} -1} & {\color{posgreen} -3} & {\color{negred} +0} & {\color{negred} +0} & {\color{posgreen} -1} & {\color{negred} +0} \\
\textbf{Llama 70b} & 0 / 2 & 1 / 3 & 0 / 0 & 0 / 5 & 1 / 1 & 0 / 0 \\
\textit{Delta} & {\color{posgreen} -2} & {\color{posgreen} -2} & {\color{negred} +0} & {\color{posgreen} -5} & {\color{negred} +0} & {\color{negred} +0} \\
\textbf{GPT 4.1} & 1 / 1 & 1 / 1 & 0 / 0 & 0 / 0 & 0 / 0 & 0 / 0 \\
\textit{Delta} & {\color{negred} +0} & {\color{negred} +0} & {\color{negred} +0} & {\color{posgreen} -0} & {\color{posgreen} -0} & {\color{negred} +0} \\
\bottomrule 
\end{tabular}
% }
\caption{Comparison of entity tracking accuracy under perturbations for different models on the cricket dataset. Extra (\%) denotes the proportion of spurious entities generated that are not present in the ground truth, Miss (\%) denotes the proportion of ground-truth entities omitted by the model, and Ent (\%) measures cross-role entity entanglement, defined as the percentage of predicted entities appearing in the opposing ground-truth role (batsman vs.\ bowler)}
\label{tab:cricket_extra_missing}
\end{table}

\begin{table}[h!]
\centering
\setlength{\aboverulesep}{0pt}
\setlength{\belowrulesep}{0.2pt}
\footnotesize
\setlength{\tabcolsep}{3.5pt} 
% \resizebox{\linewidth}{!}{% 
{
\begin{tabular}{l|cc}\toprule
\rowcolor{rowgray} 
\textbf{Models} & \multicolumn{2}{c}{\textbf{Error Rates}} \\\midrule

% --- SECTION 1: COUNTERFACTUAL ---
 \multicolumn{3}{c}{\cellcolor{rowgray}\textbf{Counterfactual}} \\
 \multicolumn{3}{c}{\cellcolor{rowgray}(Original / Perturbed)}\\\midrule
\rowcolor{headergray} 
& Extra\% & Miss\% \\
\textbf{Gemini 2.5} & 0 / 41 & 1 / 7 \\
\textit{Delta} & {\color{negred} +41} & {\color{negred} +6} \\
\textbf{Qwen 72b} & 1 / 33 & 3 / 15 \\
\textit{Delta} & {\color{negred} +32} & {\color{negred} +12} \\
\textbf{Llama 70b} & 1 / 40 & 1 / 9 \\
\textit{Delta} & {\color{negred} +39} & {\color{negred} +8} \\
\textbf{GPT 4.1} & 1 / 44 & 1 / 3 \\
\textit{Delta} & {\color{negred} +43} & {\color{negred} +2} \\

% --- SECTION 2: ANONYMIZED ---
\rowcolor{rowgray} 
& \multicolumn{2}{c}{\textbf{Anonymized}} \\
\rowcolor{rowgray} 
& \multicolumn{2}{c}{(Original / Perturbed)} \\
\rowcolor{headergray} 
& Extra\% & Miss\% \\
\textbf{Gemini 2.5} & 0 / 50 & 1 / 19 \\
\textit{Delta} & {\color{negred} +50} & {\color{negred} +18} \\
\textbf{Qwen 72b} & 1 / 49 & 3 / 27 \\
\textit{Delta} & {\color{negred} +48} & {\color{negred} +24} \\
\textbf{Llama 70b} & 1 / 39 & 1 / 19 \\
\textit{Delta} & {\color{negred} +38} & {\color{negred} +18} \\
\textbf{GPT 4.1} & 1 / 68 & 1 / 45 \\
\textit{Delta} & {\color{negred} +67} & {\color{negred} +44} \\ \bottomrule           
\end{tabular}}
\caption{Comparison of entity tracking accuracy un-
der perturbations for different models on the basketball
dataset. Extra (%) denotes the proportion of spurious
entities generated that are not present in the ground truth,
Miss (\%) denotes the proportion of ground-truth entities
omitted by the model}
\label{tab:nba}
\end{table}

\section{Temporal Robustness}
\label{sec:temporal_robustness}

In this section, we analyze the stability of model performance over time, specifically investigating how context length and the presence of explicit summary cues influence the ability to reconstruct scorecards across an innings.

\subsection{Impact of Summary Cues on Temporal consistency}
\label{sec: Temporal trends with summary}

\begin{figure}[h!]
  \centering
  % First subplot - Batsman
  \begin{subfigure}[b]{0.9\linewidth}
    \centering
    \includegraphics[width=\linewidth]{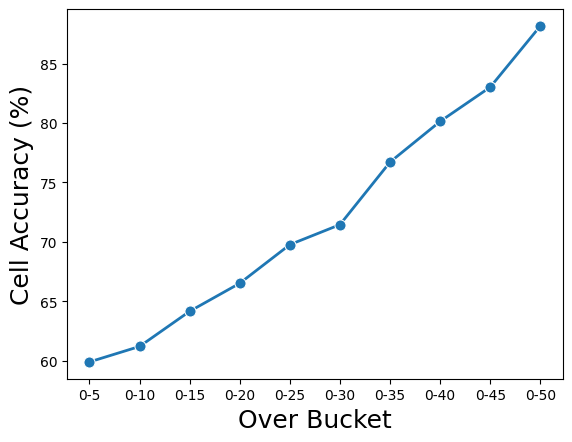}
    \caption{Batsman cell accuracy (Gemini 2.5 Flash).}
    \label{fig:temporal_with_summary_batsman}
  \end{subfigure}
  \vspace{0.5em} % space between subplots
  % Second subplot - Bowler
  \begin{subfigure}[b]{0.9\linewidth}
    \centering
    \includegraphics[width=\linewidth]{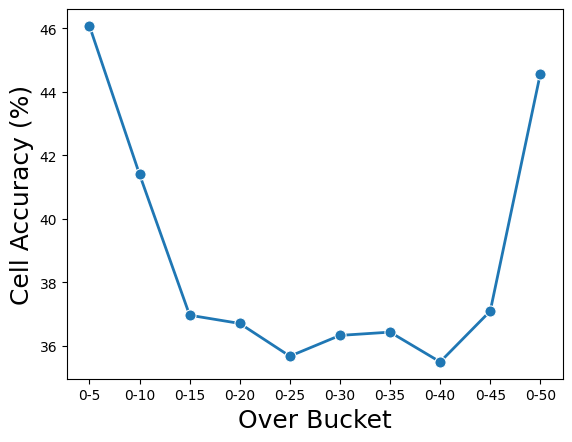}
    \caption{Bowler cell accuracy (Gemini 2.5 Flash).}
    \label{fig:temporal_with_summary_bowler}
  \end{subfigure}
  \caption{Temporal cell accuracy trends \textbf{with summary cues} for (a) batsman and (b) bowler tables. Note the rising trend in batsman accuracy.}
  \label{fig:temporal_with_summary_subplots}
\end{figure}

We have discussed how presence of cues or summaries affect models' performance in Table \ref{tab:Masked_vs_unmasked}, Figure \ref{fig:temporal_with_summary_subplots} reveals a distinct temporal trend when summary cues are present.

% --- FIGURE: TEMPORAL TRENDS (WITH SUMMARY) ---

As shown in Figure \ref{fig:temporal_with_summary_subplots}(a), batsman cell accuracy rises steadily across overs. This suggests that as wickets fall, explicit mentions of dismissals allow models to extract data directly rather than inferring it. This indicates that high performance in the presence of summaries is largely extractive rather than reasoning-based.

\subsection{Reasoning Without Cues: Temporal Trends}
\label{sec: Temporal trends without summary}

\begin{figure*}[t]
        \centering
        \begin{subfigure}[b]{\textwidth}
                \centering
                \includegraphics[width=0.85\textwidth]{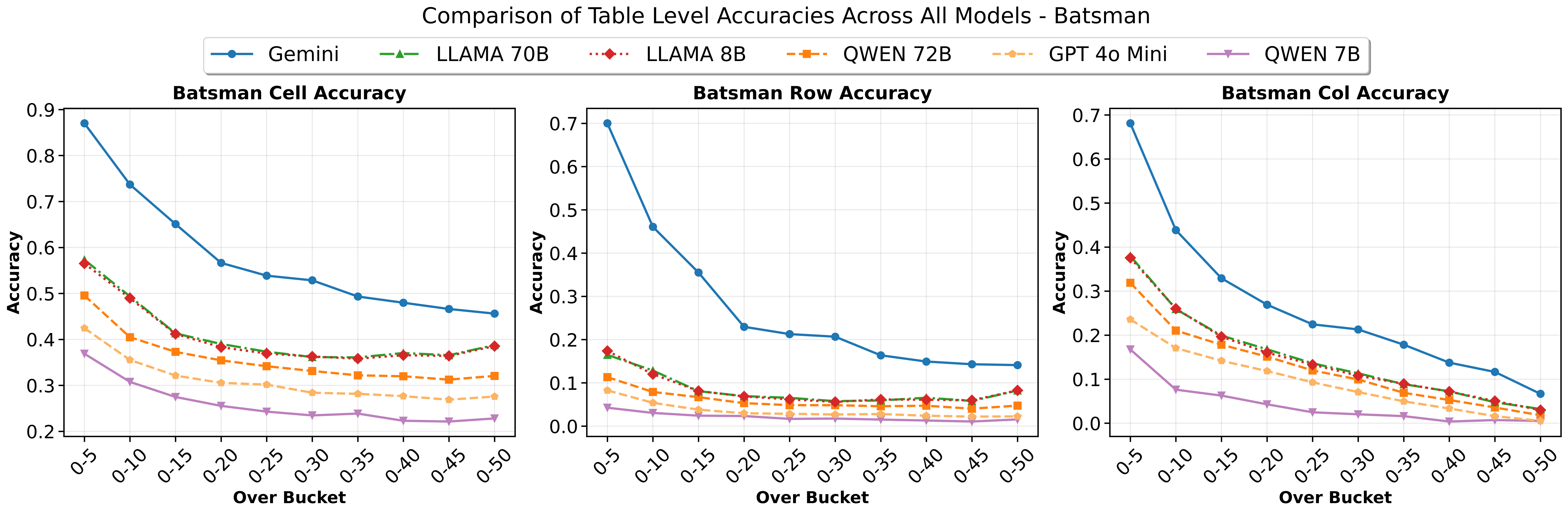}
                \label{fig:batsman}
        \end{subfigure}
        \vspace{1em}
        \begin{subfigure}[b]{\textwidth}
                \centering
                \includegraphics[width=0.85\textwidth]{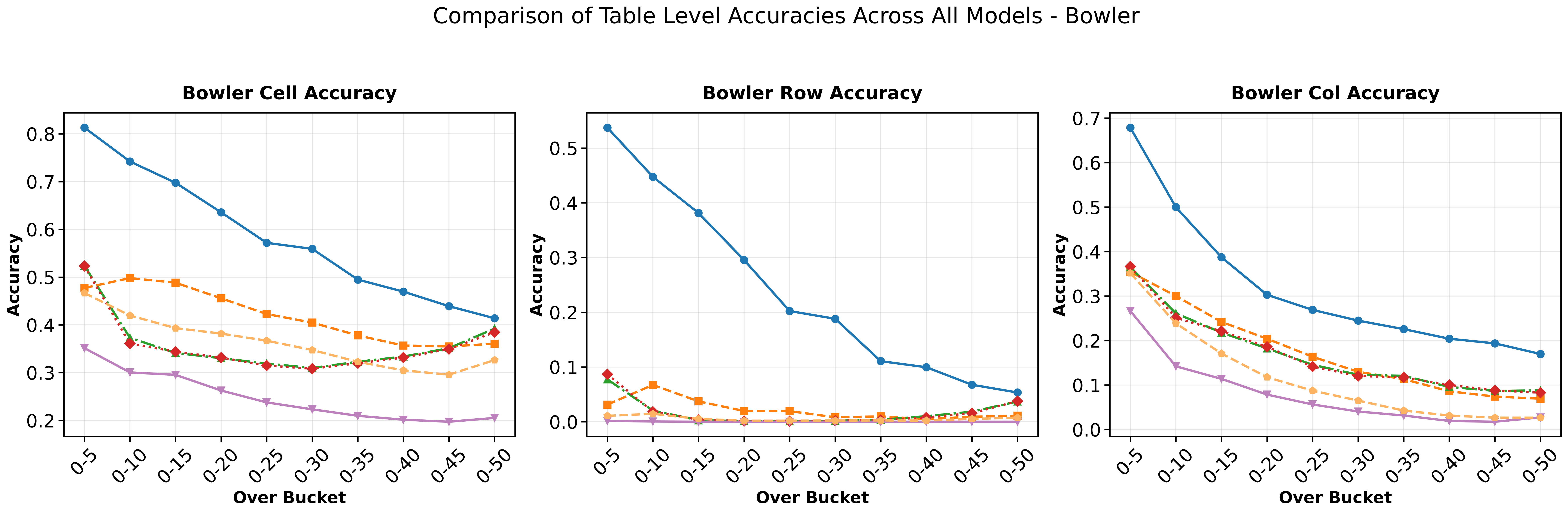}
                % \caption{
                % AUC-based Ranking for Batsman: Gemini (3.25), LLAMA-70B (1.87), LLAMA-8B (1.85), QWEN-72B (1.59), GPT 4o-Mini (1.28), QWEN-7B (0.94)}
                % \caption{
                % AUC-based Ranking for Bowler: Gemini (3.35), QWEN-72B (1.83), LLAMA-70B (1.57), LLAMA-8B (1.56), GPT 4o-Mini (1.41), QWEN-7B (0.95)}
                \label{fig:bowler}
        \end{subfigure}
        \caption{Temporal Trend of Accuracy on Cricket Score Tables}
        \label{fig:temporal_trends}
\end{figure*}

To isolate genuine temporal reasoning capabilities, we evaluated models without summary cues. Figure \ref{fig:temporal_trends} illustrates the accuracy trends over the course of an innings in this harder setting.

Unlike the cue-assisted setting, batsman accuracy without summaries declines monotonically with increasing input length (Figure \ref{fig:temporal_trends}). This confirms that without explicit state-resets (summaries), models struggle to maintain the hidden state of the game over long contexts. The decline stabilizes between overs 20–30, a "middle-overs" phase characterized by fewer events and reduced contextual drift.

Bowler accuracy curves are notably flatter. While overall accuracy is lower (consistent with the commentary's batsman-centric bias), the lack of sharp fluctuations suggests that bowling reconstruction relies more on aggregate reasoning than on specific lexical triggers.

\subsection{Quantifying Consistency: AUC Analysis}

To quantify these trends, we computed the Area Under the Curve (AUC) by averaging cell, row, and column accuracies across all overs. This metric serves as a proxy for a model's temporal stability.
% --- TABLE: AUC SCORES ---
\begin{table}[t!]
\centering
\small
\renewcommand{\arraystretch}{1.2}
\begin{tabular}{lcc}\toprule
\rowcolor[HTML]{EFEFEF} 
\textbf{Model} & \textbf{AUC (Batsman)} & \textbf{AUC (Bowler)} \\ \midrule
\textbf{Gemini 2.5 Flash} & \textbf{3.25} & \textbf{3.35} \\
\textbf{Llama (70b)} & 1.87 & 1.57 \\
\textbf{Qwen (72b)} & 1.59 & 1.83 \\
\textbf{GPT-4o-mini} & 1.28 & 1.41 \\\bottomrule
\end{tabular}
\caption{AUC scores computed from temporal accuracy curves. Higher values indicate stronger consistency in maintaining game state across long contexts.}
\label{tab:auc_trends}
\end{table}
As shown in Table \ref{tab:auc_trends}, \textbf{Gemini 2.5 Flash} demonstrates superior temporal robustness, achieving the highest AUC for both batsman and bowler tables. Notably, \textbf{Qwen 72b} outperforms Llama 70b on the bowler table, suggesting it may possess stronger aggregate reasoning capabilities despite being less effective at instantaneous extraction. The data confirms that while larger models generally maintain state better, the specific architecture (e.g., Gemini's long-context optimization) plays a critical role in temporal robustness.

\begin{table*}[h]
\centering
\definecolor{lightgray}{gray}{0.95} % specific color definition
\rowcolors{2}{lightgray}{white} % Alternating colors starting from 2nd row
\resizebox{\textwidth}{!}{%
\begin{tabular}{lrrrrrr}
\toprule
 & \multicolumn{3}{c}{\textbf{Cricket}} & \multicolumn{3}{c}{\textbf{Basketball}} \\
\cmidrule(lr){2-4} \cmidrule(lr){5-7}
\textbf{Baseline Method} & \textbf{API Calls} & \textbf{Input Tok.} & \textbf{Output Tok.} & \textbf{API Calls} & \textbf{Input Tok.} & \textbf{Output Tok.} \\
\midrule
CoT & 1 & 7,136 & 358.6 & 1 & 5,116 & 826 \\
Intermediate Sum. (L2M, n=2) & 2 & 7,975.6 & 521.2 & 3 & 6,852.7 & 1,577.3 \\
Intermediate Sum. (L2M, n=4) & 4 & 9,461.2 & 791.8 & 5 & 9,223.4 & 2,546.1 \\
Intermediate Sum. (L2M, n=8) & 8 & 12,411.3 & 1,301.2 & 9 & 13,189.4 & 3,947.5 \\
EntityCoT & 16.1 & 28,713.1 & 1,121.4 & 24.3 & 17,657.1 & 2,547.5 \\
T3 with Rule Integration & 3 & 9,756 & \textbf{7,515.5} & 4 & 8,138 & \textbf{7,813.1} \\
\bottomrule
\end{tabular}
}
\caption{Resource usage comparison across baselines for Cricket and Basketball (on sucessful samples)}
\label{tab:resource_usage}
\end{table*}

\section{Implementation Details and Practical Considerations}
\label{f_implementation details}
To ensure deterministic behavior, we use a fixed temperature of 0.1 and top-$p$ of 0.1. Llama and Qwen models are run locally on two nodes of H200 clusters, while Gemini and GPT models are accessed via Google’s and OpenAI’s official API endpoints, respectively. Beyond aggregate performance, several baselines exhibited reliability issues that materially affected feasibility, throughput, and compute cost. We summarize these behaviors to contextualize the reported results and clarify the operational trade-offs of different modeling strategies.

\paragraph{Reliability Challenges}
For Text-Tuple-Table, we observed frequent degenerate repetitive patterns in which models repeatedly emitted identical tuples hundreds or thousands of times without termination. Such outputs were unparsable and required re-generation. These failures were most common when using flat tuple formats of the form \texttt{(entity, attribute, value)}, which proved less stable for autoregressive generation than structured JSON outputs.

LLaMA~3.3 (70B) exhibited a systematic cumulative counting behavior on high-frequency attributes such as \texttt{Balls\_Faced} and \texttt{Balls}, generating running totals (e.g., 1,2,3,4,\ldots) rather than incremental updates. This led to extreme numerical errors, with RMSE values exceeding 100 on many samples. Correct outputs were obtained only after repeated regeneration, substantially increasing compute requirements.

Due to these reliability issues, multiple passes and iterative engineering of the prompts were required to successfully process the full dataset. Under Text-Tuple-Table, GPT-4.1 requires 3 passes (36/1362 (2.6\%) failure cases in first pass) to obtain valid outputs for all samples, with roughly 20\% of samples needing re-generation. LLaMA~3.3 required up to 15 passes (648/1362 (47.6\%) failure cases in the first pass) to achieve complete coverage, significantly increasing wall-clock time despite its lower per-token cost when run locally. Gemini-2.5-flash was the most effective model, with only 2\% failure cases, requiring only two retries. 

Table \ref{tab:resource_usage} details the computational footprint across all baselines, highlighting the significant variance in resource consumption required by structured reasoning frameworks. Standard CoT remains the most economical approach, serving as the efficiency baseline with a single API call and minimal token usage ($5\text{--}7$k input tokens). In contrast, EntityCoT exhibits the highest computational demand, requiring an average of $16.1$ to $24.3$ API calls per instance, an increase of over $20\times$ compared to CoT, along with the highest input token load ($28.7$k for Cricket). Meanwhile, T3 presents a unique profile: while its API call overhead is moderate ($3\text{--}4$ calls), it generates significantly higher output volumes ($~7.5$k tokens), reflecting its comprehensive table-generation process. The Intermediate Summarisation (L2M) methods demonstrate a linear scaling in cost, where increasing $n$ from 2 to 8 proportionally raises API calls and token usage without reaching the resource intensity of EntityCoT.

\paragraph{Model-Specific Feasibility Constraints.}
Qwen models were not evaluated on the most complex baselines due to extremely low throughput and persistent instability in producing structured outputs under these settings. Accordingly, tuple-based and entity-centric baselines are reported only for models capable of reliably emitting well-formed structured representations. Overall, these observations highlight that high average performance does not necessarily imply practical deployability, and that reliability, retry behavior, and structured output stability are critical factors for long-context table generation.

\section{Results Tables}
We have done our analysis on two sets of data: ODI cricket (50 over matches) and NBA basketball. The nature of data differs not only in terms of sport but also in other aspects like recency and temporal span—the basketball data is from 2021-2022, in contrast to ODI cricket which spans from 2006-2025. Additionally, the datasets differ in their structural characteristics: basketball features continuous fluid play with frequent position switches, while cricket has discrete, turn-based ball-by-ball commentary with fixed role distinctions between batsmen and bowlers.

\section{Ground Truth Generation}
\label{sec: Pseudocode}
\subsection{Structuring Cricket Commentary from CricInfo}
We convert every ball bowled into a structure using regexes to extract the batsman and bowler entities which always have a pattern. We also extract the runs that comes off that bowl to keep tracking the numerical attributed linked to bowler and batsman. The pseudocode for the same has been explained below.
% in body
\begin{lstlisting}[style=compactcode]
Input: raw_commentary_text

1) Segment into balls
   - Find over markers with regex: r"\b([0-4]?[0-9]\.[1-6])\b"
   - For each adjacent pair of matches, slice text between them → ball_chunk

2) Parse each ball_chunk
   - Split by lines → L
   - over := float(first space-token in ball_chunk)  // from L joined/split by spaces
   - run_token := normalize(L[1]); map "•"→"0"
   - Extract bowler,batter from L[2] using:
       regex: r"^(?P<bowler>.+?) to (?P<batter>.+?)," → groups["bowler"], groups["batter"]
   - commentary := concatenate L[2:] until any stop condition:
       * contains "CRR:" or "end of over"
       * contains time token via regex: r"\b\d{2}(?:am|pm)\b" (case-insensitive)
       * contains " SR: "
     (Stop at first hit; keep preceding text.)

3) Derive per-ball features
   - Team runs increment:
       if run_token starts with digit → int(run_token[0]); else 0
   - Batter runs:
       if regex r"^\d+$" → int(run_token)
       elif regex r"^\d+nb$" → int(first digit) - 1
       else 0
   - Batter 4/6:
       four := 1 if (run_token == "4" or regex r"^4(?!lb)") else 0
       six  := 1 if (run_token == "6" or regex r"^6(?!lb)") else 0
   - Batter ball faced:
       1 if (regex r"^\d+$" or run_token == "W" or (no 'w' in run_token)) else 0
   - Bowler legal ball:
       1 if (regex r"^\d+$" or (no 'w' and no 'nb' in run_token)) else 0
   - Bowler runs conceded:
       if regex r"^\d+$" → int(run_token)
       elif regex r"^\d+(?:w|nb)$" → int(first digit)
       else 0
   - Wicket (bowler credit):
       if run_token == "W" AND commentary contains any dismissal keyword:
         [" st "," c "," lbw "," hit wicket "," b "] → wicket = 1
       else 0

4) Continuity enforcement
   - Compute ball index: idx(over) = floor(over)*6 + int(10*(over - floor(over)))
   - If previous run was wide/nb (regex r"^\d+(?:w|nb)$"), expect same idx; else expect idx+1.
   - Skip chunk if expectation not met.

5) Collect row
   - Append: {raw_text, over, run_token, team_runs_inc, commentary,
              bowler, batter, batter_runs, four, six, batter_balls,
              bowler_balls, bowler_runs, wicket, dismissal_phrase}

6) Build DataFrame from all rows
\end{lstlisting}
\label{sec: example_gnd_truth}
\subsection{Commentary Ground Truth: Cricket}

A sample cricket commentary from our dataset has been shown along with the ground truth that was calculated using the pipeline explained in section \ref{sec: Pseudocode}.
\begin{lstlisting}[style=compactcode]
Muzarabani to Tamim, no run fullish ball on middle and leg, clipped to square leg

Muzarabani to Tamim, no run full and straighter on off this time. Driven down the ground with the full face of the bat

Muzarabani to Tamim, no run past the outside edge to the batter. Fullish length in the off-stump channel, as Tamim presents his bat playing in the line and the ball shoots of the the pitch

Muzarabani to Tamim, 1 wide good length ball pitched outside leg, as he shoulders arms. Wide called

Muzarabani to Tamim, no run fuller on middle and off. He goes across the stumps to clip that to square leg

Muzarabani to Tamim, no run beaten on the outside edge. Good length delivery pitched on middle and leg, and angles away as he tries to dab at that with an open face

Muzarabani to Tamim, no run good length again, this time on off. Good bounce for Muzarabani again, as Tamim blocks that off the front foot close to his body

Chatara to Litton Das, no run full and wide of off to perhaps draw the drive, and angles away further as Liton shoulders arms

Chatara to Litton Das, no run good length ball on middle and off. Defended off the back foot to cover Two slips in place for Chatara

Chatara to Litton Das, no run full ball again and closer to off. This was around fifth stump, and Liton was happy to let that go again

Chatara to Litton Das, no run full length just outside off. Liton takes a good stride forward to tap that to cover

Chatara to Litton Das, no run fullish length and slightly wide of off. Straightens after pitching after Liton leaves it alone

Chatara to Litton Das, no run another full delivery just outside off. Defended to cover off the front foot

Muzarabani to Tamim, no run full length just outside off, and tapped to cover with an open face

Muzarabani to Tamim, no run pulls his length back a touch just outside off. Driven toward mid-on with the full face of the bat

Muzarabani to Tamim, FOUR runs slashed on the up through the covers and with a crackling sound. Good length ball outside off, and Tamim takes a solid foot forward to hit that aerially and holds the pose after that too

Muzarabani to Tamim, no run good length on middle and leg, clipped to square leg off the back foot

Muzarabani to Tamim, FOUR runs glorious drive between mid-off and cover. Presents the full face of the bat to this full delivery outside off, and gets the second boundary of the over

Muzarabani to Tamim, no run fullish length outside off, and Tamim plays within the line as the ball passes his outside edge

Chatara to Litton Das, no run beats him on the outside edge. Fullish, nagging length in the channel with the ball angling away as he tries to defend at that

Chatara to Litton Das, no run full ball on off again, and defended to point this time

Chatara to Litton Das, no run goes full but much closer to off stump. Driven to mid-off with the full face of the bat

Chatara to Litton Das, 1 wide fullish ball pitching on middle and leg, and angling down leg further. Liton shoulders arms, and it's called a wide

Chatara to Litton Das, no run another fullish ball, and straighter as well. Pushed off the front foot to cover

Chatara to Litton Das, FOUR runs flicked off his pads to deep square leg, as the bowler goes full on leg this time. Just clips that to the right of the square leg umpire and beats fine leg to his left

Chatara to Litton Das, no run driven off the full face to point, as Chatara bowls this on a good length and slightly wide of off

Muzarabani to Tamim, no run good carry for Muzarabani off a full length in the off-stump channel. Tamim shoulders arms to that, and the keeper collects that really high

Muzarabani to Tamim, 1 run pushes this off the outside edge with his front leg in the air as the bowler goes full and very close to off. Third man rushes to his right to dive and keep that to a single

Muzarabani to Litton Das, no run good length delivery on off, and angling in as he fends that to the on side off his chest

Muzarabani to Litton Das, no run full and on off, as Liton gets a leading edge back to the bowler while trying to flick that to the on side

Muzarabani to Litton Das, no run another full ball on off, and Liton is right forward to block that back

Muzarabani to Litton Das, no run the ball just evades his off stump as he defends this good length ball close to himself. It was angling in after pitching outside off, and Liton played that late with the ball escaping his off stump with a few bounces away

\end{lstlisting}

Tables~\ref{tab: Bowler Table Summary 5 overs} and~\ref{tab: Batsman Table Summary 5 overs} illustrate example ground-truth scorecards derived from a five-over cricket commentary segment. The \textbf{bowler table} reports the number of balls delivered, total runs conceded, wickets taken, overs bowled, and maiden overs for each bowler. The \textbf{batsman table} summarizes runs scored, balls faced, boundaries (fours and sixes), and strike rate (runs per 100 balls faced) for each batsman. These columns reflect standard cricket statistics and are deterministically aggregated from ball-level events.

\begin{table}[h!]
\footnotesize
\centering
\resizebox{\columnwidth}{!}{%
\begin{tabular}{lccccc}\toprule
\rowcolor[HTML]{C0C0C0} 
\cellcolor[HTML]{EFEFEF}Bowler & Balls & Runs & Wickets & Overs & Maidens \\\midrule
\cellcolor[HTML]{EFEFEF}Chatara    & 12 & 5  & 0 & 2.0 & 1 \\
\cellcolor[HTML]{EFEFEF}Muzarabani & 18 & 10 & 0 & 3.0 & 0 \\ \bottomrule
\end{tabular}
}
\caption{Bowling summary}
\label{tab: Bowler Table Summary 5 overs}
\end{table}

\begin{table}[h!]
\resizebox{\columnwidth}{!}{%
\begin{tabular}{ l  l  l  l  l  l  }
\toprule
\rowcolor[HTML]{C0C0C0} 
\cellcolor[HTML]{EFEFEF}Batsman& \textbf{Runs} & \textbf{Balls\_Faced} & \textbf{Fours} & \textbf{Sixes} & \textbf{S/R} \\
\midrule
\cellcolor[HTML]{EFEFEF}Litton Das& 4.0 & 16 & 1 & 0 & 25.0 \\

\cellcolor[HTML]{EFEFEF}Tamim& 9.0 & 14 & 2 & 0 & 64.29 \\
\bottomrule

\end{tabular}
}
\caption{Batsman Summary}
\label{tab: Batsman Table Summary 5 overs}
\end{table}

\subsection{Structuring Basketball Commentary}
\label{sec: Pseudocode_Basketball}

The flow below explains how Basketball commentaries are collected from ESPN.
\begin{lstlisting}[style=compactcode]
Input: ESPN_play_by_play_pages

1) Scrape play-by-play data
   - Initialize thread pool for concurrent Selenium sessions
   - For each game URL:
       * Launch headless browser
       * Navigate to ESPN Deportes play-by-play page
       * Scroll/load dynamic content
       * Extract event elements (quarter, clock, description, score)
       * Handle timeouts and retry failures
   - Aggregate all events into raw JSON structure keyed by game_id

2) Parse raw JSON into structured commentary
   - For each game_id in JSON:
       * Group events by quarter
       * Within each quarter, sort by game clock (descending)
       * For each event:
           - Extract timestamp, event text, score snapshot
       * Concatenate events into chronological narrative blocks
   - Write one text file per game with quarter headers and timestamps

3) Clean commentary text
   - Iterate through generated text files
   - Apply regex filters to remove score identifiers:
       * Remove patterns like r"\(\d+-\d+\)"
       * Remove inline score tokens (e.g., "LAL 102-98")
   - Normalize whitespace and line breaks
   - Preserve only descriptive play-by-play language

4) Output refined corpus
   - Save cleaned files into a parallel directory structure
   - Ensure one narrative-only text file per game
   - Output optimized textual dataset for LLM ingestion

Output: clean_narrative_commentary_corpus
\end{lstlisting}

\subsection{Commentary Ground Truth: Basketball}
A sample basketball commentary from our dataset has been shown along with the ground truth that was calculated using the pipeline explained in section \ref{sec: Pseudocode_Basketball}.
\begin{lstlisting}[style=compactcode]
Nic Claxton vs. Brook Lopez (Giannis Antetokounmpo gains possession)
Grayson Allen misses 27-foot three point jumper
Kevin Durant defensive rebound
Giannis Antetokounmpo shooting foul
Nic Claxton misses free throw 1 of 2
Nets offensive team rebound
Nic Claxton misses free throw 2 of 2
Giannis Antetokounmpo defensive rebound
Giannis Antetokounmpo misses three point pullup jump shot
Brook Lopez offensive rebound
Giannis Antetokounmpo misses pullup jump shot
Blake Griffin defensive rebound
Nic Claxton makes alley oop dunk shot (James Harden assists)
Giannis Antetokounmpo misses 11-foot hook shot
Giannis Antetokounmpo offensive rebound
Giannis Antetokounmpo makes two point shot
James Harden misses 28-foot step back jumpshot
Giannis Antetokounmpo defensive rebound
Brook Lopez makes 28-foot three point jumper (Giannis Antetokounmpo assists)
Kevin Durant makes 13-foot two point shot (James Harden assists)
Giannis Antetokounmpo misses layup
Giannis Antetokounmpo offensive rebound
James Harden blocks Giannis Antetokounmpo's two point shot
\end{lstlisting}
\begin{table}[h]
\scriptsize % 1. Small but readable font for narrow columns
\centering
\setlength{\tabcolsep}{2.2pt} % 2. Tightens space between columns
\renewcommand{\arraystretch}{1.2} % 3. Adds vertical breathing room

\begin{tabular}{@{} l cccccccccc @{}} % @{} removes side margins
\toprule

% HEADER
\rowcolor[HTML]{003366}
\color{white}\textbf{Player} & 
\color{white}\textbf{PTS} & 
\color{white}\textbf{REB} & 
\color{white}\textbf{AST} & 
\color{white}\textbf{STL} & 
\color{white}\textbf{BLK} & 
\color{white}\textbf{TO} & 
\color{white}\textbf{FGM} & 
\color{white}\textbf{FGA} & 
\color{white}\textbf{3PM} & 
\color{white}\textbf{3PA} \\ \midrule

% NETS
\rowcolor[HTML]{EFEFEF}
\multicolumn{11}{l}{\textbf{\color[HTML]{003366} Brooklyn Nets}} \\ 
N. Claxton  & 2 & 0 & 0 & 0 & 0 & 0 & 1 & 1 & 0 & 0 \\
K. Durant   & 2 & 1 & 0 & 0 & 0 & 0 & 1 & 1 & 0 & 0 \\
B. Griffin  & 0 & 1 & 0 & 0 & 0 & 0 & 0 & 0 & 0 & 0 \\ 
J. Harden   & 0 & 0 & 2 & 0 & 1 & 0 & 0 & 1 & 0 & 1 \\ \midrule

% BUCKS
\rowcolor[HTML]{EFEFEF}
\multicolumn{11}{l}{\textbf{\color[HTML]{003366} Milwaukee Bucks}} \\ 
Giannis A.  & 2 & 4 & 1 & 0 & 0 & 0 & 1 & 6 & 0 & 1 \\
B. Lopez    & 3 & 1 & 0 & 0 & 0 & 0 & 1 & 1 & 1 & 1 \\
G. Allen    & 0 & 0 & 0 & 0 & 0 & 0 & 0 & 1 & 0 & 1 \\ 
\bottomrule

\end{tabular}
\caption{Q1 Summary (Compact View)}
\label{tab: Basketball sample}
\end{table}

\section{Prompts}
\label{prompt}
\subsection{Chain of Thought (COT)}
\subsection*{Prompt for Table Generation: Cricket}
The prompt used for the generating the Batsman and Bowler tables using Zero-Shot (ZS) Chain of Thought (COT) has been shown below.  We use ZS COT as it has been shown in \cite{deng-etal-2024-text} that models improve in ZS setting after applying COT. It has also been shown by \cite{deng-etal-2024-text} that few-shot learning does not improve performance and ZS performs far better than fine-tuning setting.
\paragraph{Prompt Card.}
\paragraph{System message}
\begin{promptbox}{System Prompt}
\textbf{Role:} You are an expert in building statistical summary tables from live commentary text. Return ONLY the two JSON objects according to the rules given below - no prose, no code, no markdown.

\vspace{0.5em}
\textbf{Output Rules:}
\begin{itemize}[leftmargin=1.5em, label=\textbullet, nosep]
    \item Two lines total: line 1 = batsman JSON; line 2 = bowler JSON.
    \item Use EXACT key casing: \\
    \jsonkey{batsman}, \jsonkey{Runs}, \jsonkey{Balls\_Faced}, \jsonkey{Fours}, \jsonkey{Sixes}, \jsonkey{S/R}, \jsonkey{dismissal}; \\
    \jsonkey{bowler}, \jsonkey{Balls}, \jsonkey{Runs\_Given}, \jsonkey{Wickets}, \jsonkey{Overs}, \jsonkey{Maidens}.
    \item Arrays align by index (e.g., batsman[i] $\leftrightarrow$ Runs[i] $\leftrightarrow$ Balls\_Faced[i]).
    \item Types: integers for count fields; \jsonkey{S/R} is a float with two decimals; \jsonkey{Overs} is "O.B" string (e.g., "3.4").
    \item Unknown values: use 0 (numbers) or "" (strings). Never output null/NaN.
\end{itemize}

\vspace{0.5em}
\textbf{Cricket Rules:}

\textbf{[Batsman]}
\begin{itemize}[leftmargin=1.5em, label=-, nosep]
    \item \jsonkey{Runs}: credit only off-the-bat (exclude byes/leg-byes). For no-balls, bat runs are credited separately; the +1 nb is NOT bat runs.
    \item \jsonkey{Balls\_Faced}: increment for every delivery except wides (no-balls DO count as a ball faced).
    \item \jsonkey{Fours}/\jsonkey{Sixes}: set only for off-the-bat boundaries; exclude b/lb boundaries.
    \item \jsonkey{S/R}: (Runs / Balls\_Faced) * 100, rounded to 2 decimals.
    \item \jsonkey{dismissal}: "" if not out; else short phrase such as "c Fielder b Bowler", "lbw", "run out", "st", "b Bowler", "hit wicket".
\end{itemize}

\vspace{0.5em}
\textbf{[Bowler]}
\begin{itemize}[leftmargin=1.5em, label=-, nosep]
    \item \jsonkey{Balls}: count only legal deliveries (wides/no-balls do NOT add to Balls).
    \item \jsonkey{Runs\_Given}: includes all conceded (incl. wides, no-balls, byes/leg-byes).
    \item \jsonkey{Wickets}: credit only bowler-attributable dismissals (b, c off bowler, lbw, st off bowler, hit wicket). Do NOT credit run out.
    \item \jsonkey{Overs}: floor(Balls/6) "." (Balls \% 6).
    \item \jsonkey{Maidens}: count overs with zero \jsonkey{Runs\_Given} in that over (0 if not inferable).
\end{itemize}

\vspace{0.5em}
\textbf{Validation:}
\begin{itemize}[leftmargin=1.5em, label=\textbullet, nosep]
    \item Output must be valid JSON with no trailing commas and no extra lines.
    \item Arrays must be equal length per table.
\end{itemize}
\end{promptbox}
\paragraph{User message.}
\begin{promptbox}{User Message: Data Injection \& CoT Trigger}
\textbf{Instruction:} You are about to be given a cricket commentary. Extract the final batsman and bowler scorecards by aggregating across the commentary. Return ONLY the two JSON objects as specified.

\vspace{0.5em}
\textbf{Commentary:} 
\begin{tcolorbox}[colback=white, colframe=gray!20, boxrule=0.5pt, sharp corners, left=2pt, right=2pt, top=2pt, bottom=2pt]
\texttt{\{commentary\}}
\end{tcolorbox}

\vspace{0.5em}
\textbf{Output format} (copy keys verbatim; replace placeholders with values):

\begin{itemize}[leftmargin=0pt, label={}, nosep]
\item \texttt{\{}
    \begin{itemize}[leftmargin=1.5em, label={}, nosep]
    \item \jsonkey{"batsman"}:\texttt{[batsman1, batsman2, ...],}
    \item \jsonkey{"Runs"}:\texttt{[runs1, runs2, ...],}
    \item \jsonkey{"Balls\_Faced"}:\texttt{[balls1, balls2, ...],}
    \item \jsonkey{"Fours"}:\texttt{[fours1, fours2, ...],}
    \item \jsonkey{"Sixes"}:\texttt{[sixes1, sixes2, ...],}
    \item \jsonkey{"S/R"}:\texttt{[sr1, sr2, ...],}
    \item \jsonkey{"dismissal"}:\texttt{[dis1, dis2, ...]}
    \end{itemize}
\item \texttt{\}}
\item \texttt{\{}
    \begin{itemize}[leftmargin=1.5em, label={}, nosep]
    \item \jsonkey{"bowler"}:\texttt{[bowler1, bowler2, ...],}
    \item \jsonkey{"Balls"}:\texttt{[balls\_b1, balls\_b2, ...],}
    \item \jsonkey{"Runs\_Given"}:\texttt{[runs\_g1, runs\_g2, ...],}
    \item \jsonkey{"Wickets"}:\texttt{[wkts1, wkts2, ...],}
    \item \jsonkey{"Overs"}:\texttt{[overs1, overs2, ...],}
    \item \jsonkey{"Maidens"}:\texttt{[m1, m2, ...]}
    \end{itemize}
\item \texttt{\}}
\end{itemize}

\vspace{0.8em}
\hrule
\vspace{0.5em}
% Highlighting the Chain-of-Thought Trigger
\textbf{Magic Phrase:} \texttt{\textbf{Let's think step by step.}}
\end{promptbox}
\subsection*{Prompt for Table Generation: Basketball}
\paragraph{Prompt Card.}
\paragraph{System message}
\begin{promptbox}{System Prompt: Basketball (CoT)}
\textbf{Role:} You are an expert in building statistical summary tables from live basketball play-by-play commentary text. Return ONLY one valid JSON object, following the rules below. No prose, no explanations, no markdown, no code blocks.

\vspace{0.5em}
\textbf{Output Rules:}
\begin{itemize}[leftmargin=1.5em, label=\textbullet, nosep]
    \item One line total: a single JSON object.
    \item Exact key casing (lowercase only): \\
    \jsonkey{player}, \jsonkey{pts}, \jsonkey{reb}, \jsonkey{ast}, \jsonkey{stl}, \jsonkey{blk}, \jsonkey{to}, \jsonkey{fg\_made}, \jsonkey{fg\_attempted}, \jsonkey{3pt\_made}, \jsonkey{3pt\_attempted}.
    \item Arrays must align by index (e.g., player[i] $\leftrightarrow$ pts[i] $\leftrightarrow$ reb[i] \dots).
    \item All stat values must be integers. Unknown/non-inferable values must be 0.
    \item Never output null, NaN, missing keys, or extra keys. Do not invent players.
\end{itemize}

\vspace{0.5em}
\textbf{Basketball Stat Definitions:}
\begin{itemize}[leftmargin=1.5em, label=-, nosep]
    \item \jsonkey{pts} (Points): +2 for made 2pt FG; +3 for made 3pt FG; +1 for made FT. (Do NOT award for misses/fouls).
    \item \jsonkey{fg\_made}: +1 for any made FG (2pt or 3pt). Exclude FTs.
    \item \jsonkey{fg\_attempted}: +1 for any FG attempt (made or missed). Exclude FTs.
    \item \jsonkey{3pt\_made}: +1 for a made 3-point shot.
    \item \jsonkey{3pt\_attempted}: +1 for any 3-point shot attempt (made or missed).
    \item \jsonkey{reb} (Rebounds): Count offensive and defensive. (+1 per explicit credit).
    \item \jsonkey{ast} (Assists): Credit ONLY when explicitly stated (e.g., "assisted by X"). Do not infer.
    \item \jsonkey{stl} (Steals): Credit ONLY on explicit terms (steal, strip, interception).
    \item \jsonkey{blk} (Blocks): Credit ONLY on explicit "blocked shot" statements.
    \item \jsonkey{to} (Turnovers): Credit ONLY when explicitly attributed to a player.
\end{itemize}

\vspace{0.5em}
\textbf{General Rules:}
\begin{itemize}[leftmargin=1.5em, label=\textbullet, nosep]
    \item If a player is mentioned but no stats are inferable, include player with all zeros.
    \item Aggregate stats across the entire commentary.
    \item Output must be valid JSON with no trailing commas and no extra lines.
\end{itemize}
\end{promptbox}
\paragraph{User message.}
\begin{promptbox}{User Message: Basketball Data Injection}
\textbf{Instruction:} You are about to be given basketball commentary. Extract the final per-player boxscore by aggregating across the commentary. Return ONLY the single JSON object described above.

\vspace{0.5em}
\textbf{Commentary:}
\begin{tcolorbox}[colback=white, colframe=gray!20, boxrule=0.5pt, sharp corners, left=2pt, right=2pt, top=2pt, bottom=2pt]
\texttt{\{commentary\}}
\end{tcolorbox}

\vspace{0.5em}
\textbf{OUTPUT FORMAT (COPY EXACTLY):}
\begin{itemize}[leftmargin=0pt, label={}, nosep]
\item \texttt{\{}
    \begin{itemize}[leftmargin=1.5em, label={}, nosep]
    \item \jsonkey{"player"}:\texttt{[p1, p2, ...],}
    \item \jsonkey{"pts"}:\texttt{[...],}
    \item \jsonkey{"reb"}:\texttt{[...],}
    \item \jsonkey{"ast"}:\texttt{[...],}
    \item \jsonkey{"stl"}:\texttt{[...],}
    \item \jsonkey{"blk"}:\texttt{[...],}
    \item \jsonkey{"to"}:\texttt{[...],}
    \item \jsonkey{"fg\_made"}:\texttt{[...],}
    \item \jsonkey{"fg\_attempted"}:\texttt{[...],}
    \item \jsonkey{"3pt\_made"}:\texttt{[...],}
    \item \jsonkey{"3pt\_attempted"}:\texttt{[...]}
    \end{itemize}
\item \texttt{\}}
\end{itemize}
\end{promptbox}
% \begin{lstlisting}[language={},basicstyle=\ttfamily\small,breaklines=true]
% OUTPUT FORMAT (COPY EXACTLY)
% {"player":[p1,p2,...],
% "pts":[...],
% "reb":[...],
% "ast":[...],
% "stl":[...],
% "blk":[...],
% "to":[...],
% "fg_made":[...],
% "fg_attempted":[...],
% "3pt_made":[...],
% "3pt_attempted":[...]}
% You are about to be given basketball commentary.
% Extract the final per-player boxscore by aggregating across the commentary.
% Return ONLY the single JSON object described above.
% \end{lstlisting}
\subsection{Entity Chain of Thought (Entity-COT)}
\subsection*{Entity-COT for Basketball}
% --- STEP 1: EXTRACTION ---
\begin{methodbox}{method_purple}{Step 1: Entity Extraction Prompt}
\textbf{Role:} You are an expert at extracting all player names from basketball play-by-play commentary.

\textbf{Instruction:} Given the following commentary, return a STRICTLY valid JSON object with exactly one field:
\begin{itemize}[leftmargin=1.5em, nosep]
    \item \jsonkey{"players"}: a list of unique full player names (strings)
\end{itemize}

\vspace{0.5em}
\textbf{Rules:}
\begin{itemize}[leftmargin=1.5em, nosep]
    \item Return ONLY valid JSON, no prose.
    \item Use full names as they appear in commentary.
    \item Deduplicate and normalize spacing.
\end{itemize}

\vspace{0.5em}
\textbf{Commentary:} \texttt{\{commentary\}}
\end{methodbox}

\vspace{1em} % Space between steps

% --- STEP 2: GENERATION ---
\begin{methodbox}{method_purple}{Step 2: Single-Player Stat Generation}
\textbf{System Prompt:} You are an expert basketball statistician.

\textbf{User Prompt:} Given this game commentary, provide ONLY the final, fully-aggregated box score for player \textbf{"\{player\}"} as a JSON object with these keys:
\texttt{["player", "pts", "reb", "ast", "stl", "blk", "to", "fg\_made", "fg\_attempted", "3pt\_made", "3pt\_attempted"]}

\vspace{0.5em}
\textbf{Basketball Rules:}
\begin{itemize}[leftmargin=1.5em, nosep]
    \item \jsonkey{pts}: total points scored
    \item \jsonkey{reb}: total rebounds
    \item \jsonkey{ast}: total assists
    \item \jsonkey{stl}: total steals
    \item \jsonkey{blk}: total blocks
    \item \jsonkey{to}: total turnovers
    \item \jsonkey{fg\_made}: total field goals made (includes 2pt and 3pt)
    \item \jsonkey{fg\_attempted}: total field goal attempts
    \item \jsonkey{3pt\_made}: total three-pointers made
    \item \jsonkey{3pt\_attempted}: total three-point attempts
\end{itemize}

\vspace{0.5em}
\textbf{Constraint:} Output only the JSON for this player. Do not invent data. Use 0 if not available.

\vspace{0.5em}
\textbf{Example Output:}
\begin{tcolorbox}[colback=white, colframe=gray!30, boxrule=0.5pt, sharp corners]
{\footnotesize
\begin{verbatim}
{
  "player": "LeBron James",
  "pts": 28, "reb": 8, "ast": 7,
  "stl": 2, "blk": 1, "to": 3,
  "fg_made": 10, "fg_attempted": 20,
  "3pt_made": 3, "3pt_attempted": 8
}
\end{verbatim}
}
% \begin{verbatim}
% {
%   "player": "LeBron James",
%   "pts": 28, "reb": 8, "ast": 7, "stl": 2, 
%   "blk": 1, "to": 3, "fg_made": 10, 
%   "fg_attempted": 20, "3pt_made": 3, 
%   "3pt_attempted": 8
% }
% \end{verbatim}
\end{tcolorbox}
\end{methodbox}
% \begin{lstlisting}[language={},basicstyle=\ttfamily\small,breaklines=true]
% You are an expert at extracting all player names from basketball play-by-play commentary.
% Given the following commentary, return a STRICTLY valid JSON object with exactly one field:
%  - "players": a list of unique full player names (strings)

% Rules:
% - Return ONLY valid JSON, no prose
% - Use full names as they appear in commentary
% - Deduplicate and normalize spacing

% Commentary:
% {commentary}

% PLAYER_STAT_PROMPT = You are an expert basketball statistician.
% Given this game commentary, provide ONLY the final, fully-aggregated box score for player "{player}" as a JSON object with these keys:
% ["player", "pts", "reb", "ast", "stl", "blk", "to", "fg_made", "fg_attempted", "3pt_made", "3pt_attempted"]

% Basketball rules:
% - pts: total points scored
% - reb: total rebounds
% - ast: total assists
% - stl: total steals
% - blk: total blocks
% - to: total turnovers
% - fg_made: total field goals made (includes 2pt and 3pt)
% - fg_attempted: total field goal attempts
% - 3pt_made: total three-pointers made
% - 3pt_attempted: total three-point attempts

% Commentary:
% {commentary}
% (Output only the JSON for this player. Do not invent data. Use 0 if not available.)
% Example output:
% {{
%   "player": "LeBron James",
%   "pts": 28, "reb": 8, "ast": 7, "stl": 2, "blk": 1, "to": 3,
%   "fg_made": 10, "fg_attempted": 20, "3pt_made": 3, "3pt_attempted": 8
% }}
% \end{lstlisting}
\subsection*{Entity-COT for Cricket}
% --- STEP 1: EXTRACTION ---
\begin{methodbox}{method_purple}{Step 1: Entity Extraction Prompt}
\textbf{Role:} You are an expert at extracting all batsman and bowler names from cricket commentary.

\textbf{Instruction:} Given the following commentary, return a STRICTLY valid JSON object with exactly two fields:
\begin{itemize}[leftmargin=1.5em, nosep]
    \item \jsonkey{"batsmen"}: a list of unique full batsman names (strings)
    \item \jsonkey{"bowlers"}: a list of unique full bowler names (strings)
\end{itemize}

\vspace{0.5em}
\textbf{Rules:}
\begin{itemize}[leftmargin=1.5em, nosep]
    \item Return ONLY valid JSON, no prose.
    \item Use full names as they appear in commentary.
    \item Deduplicate and normalize spacing.
\end{itemize}

\vspace{0.5em}
\textbf{Commentary:} \texttt{\{commentary\}}
\end{methodbox}
\vspace{0.5em}
% --- STEP 2: BATSMAN STATS ---
\begin{methodbox}{method_purple}{Step 2: Single-Batsman Stat Generation}
\textbf{Role:} You are an expert cricket statistician.

\textbf{Instruction:} Given this match commentary, provide ONLY the final, fully-aggregated batting scorecard for batsman \textbf{"\{batsman\}"} as a JSON object with these keys:
\texttt{["batsman", "Runs", "Balls\_Faced", "Fours", "Sixes", "S/R", "dismissal"]}

\vspace{0.5em}
\textbf{Cricket Rules:}
\begin{itemize}[leftmargin=1.5em, nosep]
    \item \jsonkey{Runs}: Credit only off-the-bat (exclude byes/leg-byes). For no-balls, bat runs are credited separately (+1 nb is NOT bat runs).
    \item \jsonkey{Balls\_Faced}: Increment for every delivery except wides (no-balls DO count).
    \item \jsonkey{Fours}/\jsonkey{Sixes}: Off-the-bat boundaries only.
    \item \jsonkey{S/R}: (Runs/Balls\_Faced) * 100, rounded to 2 decimals.
    \item \jsonkey{dismissal}: "" if not out; else phrase ("c Root b Anderson", "lbw", etc.).
\end{itemize}

\vspace{0.5em}
\textbf{Example Output:}
\begin{tcolorbox}[colback=white, colframe=gray!30, boxrule=0.5pt, sharp corners]
\begin{verbatim}
{
  "batsman": "KL Rahul",
  "Runs": 37, "Balls_Faced": 24, 
  "Fours": 2, "Sixes": 1, 
  "S/R": 154.16, 
  "dismissal": "c Root b Anderson"
}
\end{verbatim}
\end{tcolorbox}
\end{methodbox}

\vspace{1em}

% --- STEP 3: BOWLER STATS ---
\begin{methodbox}{method_purple}{Step 3: Single-Bowler Stat Generation}
\textbf{Role:} You are an expert cricket statistician.

\textbf{Instruction:} Given this match commentary, provide ONLY the final, fully-aggregated bowling scorecard for bowler \textbf{"\{bowler\}"} as a JSON object with these keys:
\texttt{["bowler", "Balls", "Runs\_Given", "Wickets", "Overs", "Maidens"]}

\vspace{0.5em}
\textbf{Cricket Rules:}
\begin{itemize}[leftmargin=1.5em, nosep]
    \item \jsonkey{Balls}: Count only legal deliveries (wides/no-balls do NOT add).
    \item \jsonkey{Runs\_Given}: All conceded (incl. wides, no-balls, byes/lbs).
    \item \jsonkey{Wickets}: Bowler-attributable only (b, c, lbw, st, hit wicket). No run outs.
    \item \jsonkey{Overs}: floor(Balls/6) "." (Balls \% 6).
    \item \jsonkey{Maidens}: Overs with zero \jsonkey{Runs\_Given}.
\end{itemize}

\vspace{0.5em}
\textbf{Example Output:}
\begin{tcolorbox}[colback=white, colframe=gray!30, boxrule=0.5pt, sharp corners]
\begin{verbatim}
{
  "bowler": "Anderson",
  "Balls": 36, "Runs_Given": 55, 
  "Wickets": 2, "Overs": "6.0", 
  "Maidens": 1
}
\end{verbatim}
\end{tcolorbox}
\end{methodbox}
\subsection{Text-Tuple-Table (T$^3$)}
\subsection*{T$^3$ Prompt for Cricket}
% --- SHARED INSTRUCTION ---
\begin{methodbox}{method_emerald}{Shared Component: Domain Instruction}
\textit{(This text is injected into the \{instruction\} slot for both steps below)}

\textbf{Task:} Summarise live commentary into two tables: Batsman Scorecard \& Bowler Scorecard.

\vspace{0.5em}
\textbf{Cricket Logic Definition:}
\begin{itemize}[leftmargin=1.5em, nosep]
    \item \jsonkey{Runs (Batsman)}: Credit off-the-bat only. No byes/leg-byes.
    \item \jsonkey{Balls\_Faced}: Count all deliveries except wides. (No-balls count).
    \item \jsonkey{S/R}: $(Runs / Balls\_Faced) * 100$ (rounded to 2 decimals).
    \item \jsonkey{Runs\_Given (Bowler)}: Count ALL conceded (wides, NBs, byes, leg-byes).
    \item \jsonkey{Maidens}: Count complete overs (6 legal balls) with zero \jsonkey{Runs\_Given}.
    \item \dots \textit{[Remaining field definitions omitted for brevity]}
\end{itemize}
\end{methodbox}

\vspace{1em}

% --- STEP 1: EXTRACTION ---
\begin{methodbox}{method_emerald}{Step 1: Event-to-Tuple Extraction}
\textbf{Prompt Template:}
\texttt{According to the following instruction, please extract the relevant events and information in the form of tuples, structured as (Entity, Attribute, Value).}

\vspace{0.5em}
\textbf{Extraction Guidelines:}
\begin{itemize}[leftmargin=1.5em, nosep]
    \item Format: \texttt{(Entity, Attribute, Value)}
    \item Atomic: Each tuple represents ONE specific stat update.
    \item Multiplicity: One commentary line $\rightarrow$ Multiple Tuples.
\end{itemize}

\vspace{0.5em}
\textbf{Few-Shot Examples (Enforced Output Style):}
\begin{tcolorbox}[colback=gray!5, colframe=gray!30, boxrule=0.5pt, sharp corners]
\small\ttfamily
Input: "Muzarabani to Tamim, no run"\\
Output:\\
(Tamim, Balls\_Faced, 1)\\
(Muzarabani, Balls, 1)\\
(Muzarabani, Runs\_Given, 0)

\vspace{0.3em}
Input: "Muzarabani to Tamim, FOUR runs"\\
Output:\\
(Tamim, Balls\_Faced, 1)\\
(Tamim, Runs, 4)\\
(Tamim, Fours, 1)\\
(Muzarabani, Balls, 1)\\
(Muzarabani, Runs\_Given, 4)
\end{tcolorbox}

\vspace{0.5em}
\textbf{Input:} \texttt{<Text> \{commentary\} </Text>}
\end{methodbox}

\vspace{1em}

% --- STEP 2: AGGREGATION ---
\begin{methodbox}{method_emerald}{Step 2: Symbolic Aggregation}
\textbf{Instruction:} Aggregate these tuples into the final output tables.
\begin{itemize}[leftmargin=1.5em, nosep]
    \item Group tuples by \textbf{Entity}.
    \item Sum all numeric \textbf{Attributes}.
    \item Apply derived calculations (e.g., Strike Rate).
\end{itemize}

\vspace{0.5em}
\textbf{Input Payload:}
\texttt{<Tuples> \{generated\_tuples\_from\_step\_1\} </Tuples>}

\vspace{0.5em}
\textbf{Required Output Format:}
\begin{itemize}[leftmargin=1.5em, nosep]
    \item Line 1: \jsonkey{\{"batsman": [...], "Runs": [...], ...\}}
    \item Line 2: \jsonkey{\{"bowler": [...], "Balls": [...], ...\}}
\end{itemize}
\end{methodbox}
\subsection*{T$^3$ Prompt for Basketball}
% --- SHARED INSTRUCTION ---
\begin{methodbox}{method_emerald}{Shared Component: Domain Instruction}
\textbf{Task:} Summarise live basketball commentary into a single player statistics table.

\vspace{0.5em}
\textbf{Basketball Logic Definition:}
\begin{itemize}[leftmargin=1.5em, nosep]
    \item \jsonkey{pts}: Total points (FG + FT).
    \item \jsonkey{fg\_made/attempted}: Includes 2pt and 3pt.
    \item \jsonkey{3pt\_made/attempted}: Specific subset of FGs.
    \item \jsonkey{reb}: Total rebounds (Offensive + Defensive).
    \item \jsonkey{ast, stl, blk, to}: Standard counting stats.
\end{itemize}

\vspace{0.5em}
\textbf{Required Output Structure:}
\begin{itemize}[leftmargin=1.5em, nosep]
    \item Single JSON object.
    \item Arrays align by index: \texttt{player[i] $\leftrightarrow$ pts[i] $\leftrightarrow$ reb[i]}.
\end{itemize}
\end{methodbox}

\vspace{1em}

% --- STEP 1: EXTRACTION ---
\begin{methodbox}{method_emerald}{Step 1: Event-to-Tuple Extraction}
\textbf{Prompt Template:}
\texttt{According to the following instruction, please extract the relevant events and information in the form of tuples, structured as (Entity, Attribute, Value).}

\vspace{0.5em}
\textbf{Few-Shot Examples (Enforced Output Style):}
\begin{tcolorbox}[colback=gray!5, colframe=gray!30, boxrule=0.5pt, sharp corners]
\small\ttfamily
Input: "LeBron James makes a 2-point field goal"\\
Output:\\
(LeBron James, pts, 2)\\
(LeBron James, fg\_made, 1)\\
(LeBron James, fg\_attempted, 1)

\vspace{0.3em}
Input: "Stephen Curry makes a 3-pointer"\\
Output:\\
(Stephen Curry, pts, 3)\\
(Stephen Curry, fg\_made, 1)\\
(Stephen Curry, fg\_attempted, 1)\\
(Stephen Curry, 3pt\_made, 1)\\
(Stephen Curry, 3pt\_attempted, 1)

\vspace{0.3em}
Input: "Kevin Durant grabs a rebound"\\
Output: (Kevin Durant, reb, 1)
\end{tcolorbox}

\vspace{0.5em}
\textbf{Input:} \texttt{<Text> \{commentary\} </Text>}
\end{methodbox}

\vspace{1em}

% --- STEP 2: AGGREGATION ---
\begin{methodbox}{method_emerald}{Step 2: Symbolic Aggregation}
\textbf{Instruction:} Aggregate these tuples into the final output table.
\begin{itemize}[leftmargin=1.5em, nosep]
    \item Group tuples by \textbf{Entity}.
    \item Sum all numeric \textbf{Attributes}.
    \item Return ONLY the final output in the exact format below.
\end{itemize}

\vspace{0.5em}
\textbf{Input Payload:}
\texttt{<Tuples> \{generated\_tuples\_from\_step\_1\} </Tuples>}

\vspace{0.5em}
\textbf{Required Output Format:}
\begin{tcolorbox}[colback=white, colframe=gray!30, boxrule=0.5pt, sharp corners]
\begin{verbatim}
{
  "player": [p1, p2, ...],
  "pts": [...],
  "reb": [...],
  "ast": [...],
  "stl": [...],
  "blk": [...],
  "to": [...],
  "fg_made": [...],
  "fg_attempted": [...],
  "3pt_made": [...],
  "3pt_attempted": [...]
}
\end{verbatim}
\end{tcolorbox}
\end{methodbox}

\end{document}